\documentclass[conference,letterpaper]{IEEEtran}

\usepackage[utf8]{inputenc} 
\usepackage[T1]{fontenc}
\usepackage{url}
\usepackage{ifthen}
\usepackage{cite}
\usepackage[cmex10]{amsmath} %
\usepackage[ruled,vlined,linesnumbered]{algorithm2e}
\usepackage{algorithmic}
\usepackage{amsfonts}
\usepackage{bm}
\newcommand{\SP}{${}$\hspace{0.25cm}}
\usepackage{graphicx} 
\usepackage{subcaption}
\usepackage{xcolor}
\usepackage{diagbox}
\usepackage{booktabs}

\graphicspath{ {images/} }

\newcommand{\fl}{{Federated Learning}\xspace}
\newcommand{\mave}{{Mavericks}\xspace}

\newcommand{\algoname}{{FedMS}\xspace}

\begin{document}

\title{Maverick-Aware Shapley Valuation for \\ Client Selection in Federated Learning}

\author{%
  \IEEEauthorblockN{Mengwei Yang\IEEEauthorrefmark{1},
                    Ismat Jarin\IEEEauthorrefmark{1},
                    Baturalp Buyukates\IEEEauthorrefmark{3},
                    Salman Avestimehr\IEEEauthorrefmark{3},
                    Athina Markopoulou\IEEEauthorrefmark{1}}
  \IEEEauthorblockA{\IEEEauthorrefmark{1}%
                   Department of EECS and CS, University of California, Irvine, CA, USA}
  \IEEEauthorblockA{\IEEEauthorrefmark{3}%
                    Ming Hsieh Department of ECE, University of Southern California, CA, USA}
                    Emails: \{mengwey, ijarin, athina\}@uci.edu, \{buyukate, avestime\}@usc.edu
}

\maketitle

\begin{abstract}
Federated Learning (FL) allows clients to train a model collaboratively without sharing their private data. One key challenge in practical FL systems is data heterogeneity, particularly in handling clients with rare data,  also referred to as \emph{Mavericks}. These clients own one or more data classes exclusively, and the model performance becomes poor without their participation. Thus, utilizing \mave throughout training is crucial. In this paper, we first design a Maverick-aware Shapley valuation that fairly evaluates the contribution of Mavericks. The main idea is to compute the clients' Shapley values (SV) class-wise, {\em i.e.}, per label. Next, we propose \algoname, a \underline{\textbf{M}}averick-\underline{\textbf{S}}hapley client selection mechanism for FL that intelligently selects the clients that contribute the most in each round, by employing our Maverick-aware SV-based contribution score. We show that, compared to an extensive list of baselines, \algoname achieves better model performance and fairer Shapley Rewards distribution.

\end{abstract}

\section{Introduction}

As the pace of legislation on user privacy accelerates, regulations such as the General Data Protection Regulation (GDPR) \cite{voigt2017EU} and the California Consumer Privacy Act (CCPA) \cite{bukaty2019california} have been released to give users more control over their personal information. In this landscape, \fl (FL) has been proposed \cite{mcmahan2016communication} to facilitate machine learning (ML) over decentralized user data, taking the place of traditional centralized training approaches with significant privacy challenges. In FL, many clients collaboratively train a model by only transmitting their model updates instead of their private data. Despite this increased privacy notion, practical FL systems usually face the challenge of data heterogeneity. Unlike the idealistic data center environments, in FL, participating clients usually have heterogeneous data, which can easily cause poor accuracy and slow convergence. Even though many works have tackled data heterogeneity from model performance, client selection, and rewarding perspectives in FL \cite{Infocom22,CSCW22,dai2023tackling,PMLR23}, a prevalent scenario remains largely understudied: \emph{clients with rare data}. Clients providing rare and previously unseen data are crucial to the success of the trained ML models. Training on diverse data avoids the common bias in algorithms, leading to fairer and  trustworthy ML systems. %

\begin{figure} [t!]
\centering
\includegraphics[width=0.4\textwidth]{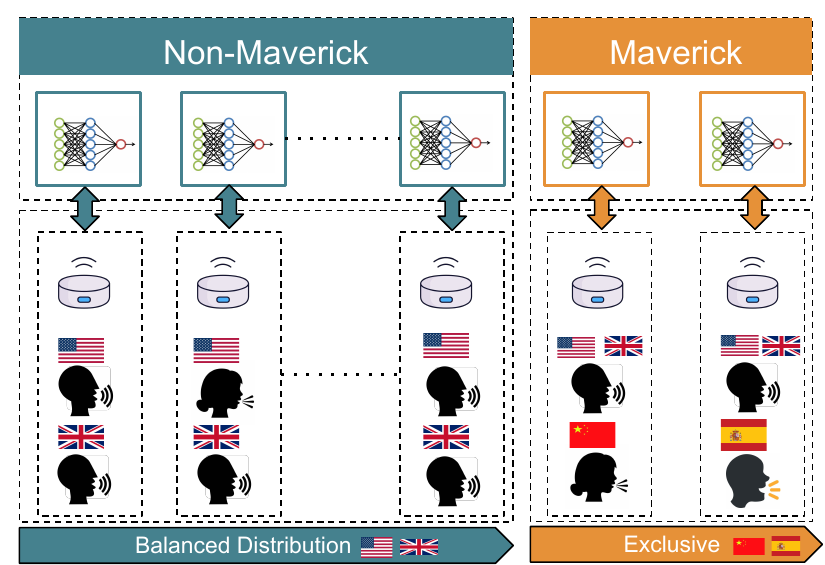}
\caption{Multiple devices participate in FL for a voice AI task. A few devices that exclusively own rare data, i.e., non-native accent data, are the Mavericks and crucial for training.}
\label{fig:maverick_illustration}
\vspace{-5mm}
\end{figure}

In \cite{huang2022tackling}, the term  \emph{\mave} was coined to refer to clients with rare data in FL, and more specifically to clients that exclusively own one or more classes ({\em i.e.}, labels) of data, whereas the non-Maverick clients have a balanced distribution from the remaining classes. 
Some examples are shown in Fig.~\ref{fig:maverick_illustration}. When training an FL model for a disease classification task, most hospitals ({\em i.e.,} clients) possess data indicating common diseases such as flu or cold. However, very few hospitals possess rare disease datasets such as for leukemia or thyroid cancers, making them Mavericks in this learning task. Another example of Mavericks is people with rare accents in training voice-activated AI systems like Amazon's Alexa and Google's Home Assistant. While the majority of these devices contain native accent data, a few of them contain data from users with non-native accents. %
 Recent studies report that these assistant devices struggle to understand non-native accents, with more than 6\% performance gap between the Western accents and the minority accents \cite{washingtonpost_alexa_accent}. This performance disparity indicates a biased performance and demonstrates the importance of training with rare (or less common) data, from  Mavericks, to create models that ``speak" to everyone.

Prior work in FL has not sufficiently addressed this problem.
The random sampling of clients at each round adopted by the conventional FL scheme, FedAvg \cite{mcmahan2017communication}, does not fully exploit rare data and can cause slow convergence, low model performance, and degraded fairness \cite{fu2023client}. Existing techniques for selecting clients in FL includes contribution-based approaches S-FedAvg \cite{nagalapatti2021game}, GreedyFed \cite{singhal2024greedy} and distance-based methods such as FedEMD \cite{huang2022tackling}. In contribution-based client selection methods, Shapley value (SV) \cite{shapley1971cores} is widely applied for measuring clients' contribution during training. Previous works \cite{huang2022tackling} and \cite{buyukates2023proof} have shown that, despite their accurate performance in i.i.d.~scenarios, SV-based methods systematically undervalue the \mave (although they are paramount for achieving high accuracy on certain classes of data), suffering from unfairness and performance loss due to under-utilization of rare data.

In this paper, we offer a principled way to value and utilize the Mavericks in FL.
We design (i) a novel Maverick-aware Shapley valuation and (ii) a corresponding client selection mechanism that can more fairly assess the contribution of \mave  and can effectively utilize them in each round. Our main contributions can be summarized as follows:
\begin{itemize}
\item We propose a class-wise SV-based contribution score to value the contributions of clients in FL. To compute this score, we define the \emph{class difficulty} in order to combine the class-wise SVs and fairly evaluate the contribution of the clients (\mave and non-\mave).
\item We then introduce \algoname, \underline{\textbf{M}}averick-\underline{\textbf{S}}hapley client selection mechanism for FL, to effectively utilize the \mave during training based on the contribution scores. We show that \algoname significantly increases the model accuracy compared to an extensive list of baselines.
\end{itemize}

\section{Problem Setup and Background}
In this section, we first formalize the FL framework \cite{mcmahan2017communication} and define \mave \cite{huang2022tackling}. We then give an overview of the SV-based methods for evaluating the contribution of clients.

\noindent\textbf{\fl (FL).} We consider a general FL system with multiple clients and one server. We let $\mathcal{K}$ denote the set of clients such that $\mathcal{K} = \{1,2,...,I\}$. Each client $i$ owns dataset $\mathcal{D}_i$, where $n_i = |\mathcal{D}_i|$. Each data point is a pair $(\bm{x}, y)$, where $\bm{x}$ is the feature vector and $y$ is the corresponding label. We let $\mathcal{M} = \{1,2,...,C\}$ denote the set of class labels. ${\bm{w}}$ is the learnable weights of the global model and each client $i$ has local model ${\bm{w}}_i$. The training objective is defined as  
\begin{align}
    \min_{\bm{w}} \mathcal{L}(\bm{w}) = \min \sum_{i \in \mathcal{K}} \frac{n_{i}}{n} \mathcal{L}_i(\bm{w}_i),
\end{align}
where we have $n = \sum_{i \in \mathcal{K}} n_i$ and the loss at client $i$ is $\mathcal{L}_i(\bm{w}_i) =  \frac{1}{n_{i}} \sum_{d \in \mathcal{D}_i} \mathcal{L}_d(\bm{w}_i)$. The FL training process includes the following steps: (i) Initialization: The server initializes the global model parameters $\bm{w}$ and broadcasts it to clients. (ii) Client Selection: In round $t$, the server selects $i \in \mathcal{K}^t = \{1,2,...,I^t\}$ clients with selection strategy $\pi$. (iii) Local Update and Model Aggregation: Each selected client $i$ in round $t$ performs local training and sends ${\bm{w}}_{i}^t$ to the server. Then, the server updates the global model using the model updates of the clients as ${\bm{w}}^{t+1} = \sum_{i =1}^{I^t} \frac{n_{i}^t}{\sum_{i =1}^{I_t} n_i^t} \bm{w}_{i}^{t}$. Steps (ii) and (iii) are repeated until the convergence of the global model.

\begin{algorithm}[th!] 
	\small
    \DontPrintSemicolon
    \SetKwFunction{UserUpdate}{UserUpdate}
    \SetKwFunction{Shapley}{Maverick-Shapley}
    \SetAlgoNoEnd
	\SetAlgoNoLine

{\bf Input:} $T$: number of training rounds; $E$: number of local epochs;
$\mathcal{K}$: set of clients; $\mathcal{D}_i$: dataset of client $i$; $B$: minibatch size; $n_i^t$: dataset size of the $i$th client in round $t$; $\mathcal{M}$: set of class labels;  $\mathcal{D}_{val}$: validation dataset; $\mathcal{V}_{class}(\cdot)$: class-wise accuracy function; $\eta_i$: learning rate at client $i$. %

	{\bf Server executes:}\\
    \SP Initialize $\bm{w}^0$, $\beta$, $\hat{S}$\\

	\SP \For{each round t = 0, ... $T-1$} {

       \textit{// Compute contribution score $\hat{S}_i$}. \\
       $\hat{S}_i = \sum \limits_{c \in \mathcal{M}} \beta^c \boldsymbol{\cdot} S_i^c, \forall i \in \mathcal{K}$\\ 
         \textit{// Sample clients from $P_{\hat{S},i}$.}\\ %
        $P_{\hat{S},i} = \frac{exp(\hat{S}_i)}{\sum \limits_{i \in \mathcal{K}}exp(\hat{S}_i)}, \forall i \in \mathcal{K}$\\
        $\mathcal{K}^{t} \gets$ sample $i$ clients $\sim P_{\hat{S},i}$\\
    
		\For{each client $i$ $\in$ $\mathcal{K}^{t}$ in parallel} {
			$\bm{w}_{i}^{t} \leftarrow$ \UserUpdate$( \bm{w}^{t}, i)$\\
			} %
          \textit{// Calculate class-wise Shapley value $\phi_i$, class difficulty $\beta$ and the best clients set $\hat{\mathcal{K}}^{t}$}.\\
			 $\phi, \beta, \hat{\mathcal{K}}^{t} \leftarrow$ \Shapley $ {\hspace{2cm}} (\{\bm{w}_i^{t}\}_{i \in  \mathcal{K}^t}, \bm{w}^{t}, D_{val}, \mathcal{V}_{class}(\cdot), \mathcal{M})$\\
        \textit{// Compute the accumulated class-wise Shapley value $S_i$}. \\
         $S_{i}^c = \alpha \boldsymbol{\cdot} S_{i}^c + (1 - \alpha) \boldsymbol{\cdot} \phi_i^c, \forall i \in \mathcal{K}^t, \forall c \in \mathcal{M}$  \\ 
         ${\bm{w}}^{t+1} \leftarrow 
        \sum \limits_{i \in \hat{\mathcal{K}}^{t}} \frac{n_{i}^t}{\sum \limits_{i \in \hat{\mathcal{K}}^{t}} n_i^t}
        \bm{w}_{i}^{t}$;\\ 
       }
	\BlankLine
	{\bf function \UserUpdate($\bm{w}^{t},i$):}\\
 	\SP  \For{each local epoch e = 1...$E$}{%
        $\mathcal{D}^B_i \gets$ select a minibatch of size $B \subseteq \mathcal{D}_i$\\
        $\bm{w}_i^{t} \gets \bm{w}_i^{t} - \eta_i \nabla \mathcal{L}_i(\mathcal{D}^B_i, \bm{w}_i^{t})$\\
	}
	\SP  \Return{${\bm{w}}_i^{t}$ to server}

\caption{\algoname: a Maverick-Shapley Client Selection Mechanism for FL
}
\label{alg:fedcsv}

\end{algorithm}

\noindent \textbf{\mave.} A Maverick is a client that owns one or more classes exclusively \cite{huang2022tackling}. Let $\mathcal{M}_{mav}$ denote the set of class labels exclusively owned by \mave. If a client is a Maverick, then its dataset satisfies $\mathcal{D}_i = \{ \{x^c, y^c\}^i_{c \in M_{mav}} ,\{x^c, y^c\}^i_{c \notin M_{mav}} \}$. Here, $\{x^c, y^c\}^i$ denotes the data points in $\mathcal{D}_i$ with label $c$. If a client is not a Maverick, then its dataset satisfies $\mathcal{D}_i = \{\{x^c, y^c\}^i_{c \notin M_{mav}} \}$. As in \cite{huang2022tackling}, we assume the data samples $\{x^c, y^c\}_{c \notin M_{mav}}$ are evenly distributed among all clients but the data samples $\{x^c, y^c\}_{c \in M_{mav}}$ are exclusively owned by the \mave. We note that there can be multiple Mavericks jointly owning the rare labels, which we call the shared \mave. 

\noindent \textbf{Shapley Value (SV) for Client Valuation in FL.} SV \cite{shapley1971cores, ghorbani2019data} of client $i$ is given by
\begin{align}
    \phi_i(\mathcal{K},\mathcal{V}) = \sum_{\mathcal{Q} \subseteq \mathcal{K}\backslash\{i\}}\frac{\mathcal{V}(\mathcal{Q}\cup\{i\}) - \mathcal{V}(\mathcal{Q})}{\tbinom{|\mathcal{K}|-1}{|\mathcal{Q}|}},
\label{eq:sv-og}
\end{align}
where $\phi_i$ is the SV for client $i$, $\mathcal{Q}$ denotes the subset of participants from $\mathcal{K}$. The utility function $\mathcal{V}(\cdot)$ can assume any form which can evaluate the utility of the input. %
The conventional SV in (\ref{eq:sv-og}) requires retraining the FL model for all subsets of clients, which is computationally prohibitive \cite{liu2022gtg}. For client contribution assessment in FL, gradient-based SV approximation techniques such as MR \cite{song2019profit}, TMR \cite{wei2020efficient}, and GTG \cite{liu2022gtg} are employed (see Appendix~\ref{app:related} for an overview). %

\section{Proposed Method: \algoname}
In this section, we describe the proposed Maverick-Shapley client selection mechanism, \algoname, that fairly computes the contributions of all clients using a class-wise SV-based contribution scoring and selects the most contributing clients in each round. The steps are outlined in Algorithm \ref{alg:fedcsv} and the list of variables is given in Appendix~\ref{app:params}.

\noindent \textbf{Maverick-Shapley Contribution Score.} 
When training a model for multi-class tasks, the difficulty of learning each class is different. Particularly in the presence of \mave, rare classes are harder to learn than the others. In order to differentiate between classes and accurately compute the contribution of each client (\mave and non-\mave alike), we propose a class-wise SV-based contribution score. In particular, we use the class-wise accuracy as the utility function in SV computations to better capture the difficulty level of each class. Class-wise accuracy is calculated as

\begin{align}\label{class_acc}
\mathcal{V}_{class}^{c}(w;\mathcal{D}_{\textrm{val}}) = \frac{N^{cc}}{\sum \limits_{j \in \mathcal{M}} N^{cj}}, \quad \forall c \in \mathcal{M},
\end{align}
where $w$ is a given model, $\mathcal{D}_{\textrm{val}}$ is validation dataset at the server, $N^{cj}$ represents the number of validation data points of class $c$ predicted as class $j$, $\mathcal{M}$ is set of class labels. Our main distinction in SV computation is the fact that we compute it in a class-wise manner to better capture the diverse resources of \mave (hence the name \emph{Maverick-Shapley}). 

In each FL round, after receiving model updates from the participating clients, the server computes the SV of a client $i$ for class $c$, $\phi_i^c$, by utilizing a gradient-based SV approximation method of its choice using (\ref{class_acc}). It then computes the accumulated SVs $S_i^c$ using a decay factor $\alpha$ as  
\begin{align}
S_{i}^c = \alpha * S_{i}^c + (1 - \alpha) * \phi_i^c, \quad \forall i \in \mathcal{K}^t, \forall c \in \mathcal{M}.
\end{align}
Finally, the server computes the contribution score of each client $i$ as a weighted sum of its class-wise accumulated SVs\footnote{In the first round, the server initializes contribution scores by calculating the cosine distance between each client model and the aggregate model.}
\begin{align}
         \hat{S}_i = \sum \limits_{c \in \mathcal{M}} \beta^c \boldsymbol{\cdot} S_i^c, \quad \forall i \in \mathcal{K},
\end{align}
where $\beta$ denotes the \emph{class difficulty}, adaptively adjusting the impact of each class in the contribution scores such that
\begin{align}
\beta^c = \frac{exp\left(\frac{1- \mathcal{V}_{class}^{c}(w;\mathcal{D}_{\textrm{val}})}{T}\right)}{\sum \limits_{c \in \mathcal{M}}exp\left(\frac{1 - \mathcal{V}_{class}^{c}(w;\mathcal{D}_{\textrm{val}})}{T}\right)}, \quad \forall c \in \mathcal{M},
\end{align}
where the temperature $T$ controls the distribution. Since the difficulty of learning each class is dynamically changing, the server updates the class difficulty $\beta$ and the contribution score $\hat{S}$ in each round.

The proposed Maverick-Shapley approach is universally applicable to the existing SV approximation algorithms. Algorithm~\ref{alg:Maverick-Shapley} describes the procedure for the MR \cite{song2019profit} technique.\footnote{Another example is in Appendix~\ref{app:maverick-getg} for the GTG-Shapley \cite{liu2022gtg} technique.}

\begin{algorithm}[t]
\small
\caption{Maverick-Shapley}
\label{alg:Maverick-Shapley}
\DontPrintSemicolon
\SetAlgoNoEnd
\SetAlgoNoLine
\SetKwFunction{ModelAverage}{ModelAverage}

\textbf{Input}: Updated client models $\{\bm{w}_i^{t}\}_{i \in \mathcal{K}^t}$; current server model $\bm{w}^{t}$;  validation dataset at server $\mathcal{D}_{\textrm{val}}$; class-wise accuracy function $\mathcal{V}_{class}(\cdot)$; $\mathcal{M}$: set of class labels.\\
 \textbf{Hyperparameter:} Temperature T \\
 \textbf{Initialize:} $\phi_i=0, \forall i\in \mathcal{K}^t$\\
\For{each subset $Q$ $\subseteq$ $\mathcal{K}^{t}$} {
$\widetilde{w}_Q=\ModelAverage(\{\bm{w}_i^{t}\}_{i \in Q},\bm{w}^{t})$\\
}

\For{client $i$ $\in$ $\mathcal{K}^{t}$} {
\For{class $c$ $\in$ $\mathcal{M}$} {

$\phi_i^c = \sum \limits_{\mathcal{Q} \subseteq \mathcal{K}^{t}\backslash\{i\}}\frac{\mathcal{V}_{class}^{c}(\widetilde{w}_{\mathcal{Q}\cup\{i\}}; \mathcal{D}_{\textrm{val}}) - \mathcal{V}_{class}^{c}(\widetilde{w}_{\mathcal{Q}}; \mathcal{D}_{\textrm{val}})}{\tbinom{|\mathcal{K}^{t}|-1}{|\mathcal{Q}|}}$\\
}
}
\textit{// Find the best clients set $\hat{\mathcal{K}}^{t}$ and its class-wise accuracy $\hat{v}$}\\
$\hat{\mathcal{K}}^{t}, \hat{v} \leftarrow \operatorname*{argmax}_{Q \subseteq \mathcal{K}^{t}} \sum \limits_{c \in \mathcal{M}} \mathcal{V}_{class}^c (\widetilde{w}_Q, \mathcal{D}_{\textrm{val}})$ \\
\textit{// Obtain  class difficulty $\beta$}\\
$\beta^c = \frac{exp(\frac{1 - \hat{v}^c}{T})}{\sum \limits_{c \in M}exp(\frac{1 - \hat{v}^c}{T})}, \forall c \in M$ \\

\Return{$\phi, \beta, \hat{K}^{t}$}
\end{algorithm}

\noindent \textbf{Client Selection.} Based on the contribution scores, the server selects the most contributing clients in each FL training round. To this end, it calculates the selection probability of each client according to their contribution scores $\hat{S}$ as
\begin{align}
P_{\hat{S},i} = \frac{exp(\hat{S}_i)}{\sum \limits_{i \in \mathcal{K}}exp(\hat{S}_i)},  \quad \forall i \in \mathcal{K},
\end{align}
and samples clients based on $P_{\hat{S}}$ in each round. As \mave exclusively own certain classes, they are the most contributing clients for those rare classes and have higher probability to be selected when the model performs poorly on rare classes. 

As the server computes the class-wise accuracy considering multiple client permutations during the contribution score computation (e.g., line 8 in Algorithm~\ref{alg:Maverick-Shapley}), it can further refine the client selection. In each training round in \algoname, the server finds the subset of clients leading to the highest total class-wise accuracy increase, i.e., the \emph{best client set} $\hat{\mathcal{K}}^t$ in line 10 in Algorithm~\ref{alg:Maverick-Shapley}, and aggregates only their updates.

\noindent \textbf{Shapley Rewards (SR).} In each round, the server computes the SV of a selected client $i$ for class $c$, $\phi_i^c$. It then calculates the \emph{Shapley Rewards} of each client $i$ for round $t$ as a weighted sum of its class-wise SVs using the current class difficulty $\beta$
\begin{align}\label{Shapley_rewards}
         R_i^t = \sum \limits_{c \in \mathcal{M}} \beta^c \boldsymbol{\cdot} \phi_i^c, \quad \forall i \in \mathcal{K}^t.
\end{align}

\section{Evaluation}

In this section, we comprehensively evaluate the effectiveness of our algorithm, \algoname, on two datasets against six baselines. We demonstrate an improved accuracy and fairer Shapley rewards for both \mave and non-\mave.

\noindent \textbf{Datasets and Models.} We use two benchmark datasets, (i) MNIST\cite{deng2012mnist} consisting of handwritten digits, with 60,000 samples for training and 10,000 for testing, and (ii) CIFAR-10 \cite{krizhevsky2009cifar} consisting of colored images of 10 classes, with 50,000 samples for training and 10,000 for testing. We utilize a lightweight MLP neural network \cite{popescu2009multilayer} for MNIST and a commonly employed CNN \cite{albawi2017understanding} for the CIFAR-10 dataset.

\noindent \textbf{Implementation Details.} Both MNIST and CIFAR-10 datasets are uniformly distributed across all 10 class labels. Here, to satisfy our Mavericks setting, we split the dataset into two scenarios: (i) 5 clients (4 non-\mave and 1 Maverick) without client selection and (ii) 50 clients (48 non-\mave and 2 \mave) with 10\% selection rate of 50 clients in each round. Each Maverick exclusively owns one class in both scenarios (i and ii). The training process involves 100 global training rounds for MNIST and 200 for CIFAR-10 both with a batch size of 64; the learning rate is 0.05 in both datasets. We employ 1 local training on MNIST and 10 local training on CIFAR-10. We choose $\alpha$ as 0.6 for both MNIST and CIFAR-10. In the proposed \algoname, we use class-wise GTG-Shapley (shown in Appendix~\ref{app:maverick-getg}) for Shapley Rewards computation. We note that our class-wise approach is applicable to other SV approximation methods such as MR and TMR as well (see Tables~\ref{tab:Mnist_accuracy} and~\ref{tab:cifar10_accuracy} for performance under different methods).

\noindent \textbf{Evaluation Metrics.} To assess the effectiveness of evaluated mechanisms, we consider the test accuracy as the utility metric. %
In addition, we evaluate different schemes based on their Shapley Rewards (SR) to the \mave. A larger SR is associated with higher contributions. While we compute SR as in (\ref{Shapley_rewards}), previous works use SVs of the clients simply as SR.

\noindent \textbf{Baselines.}
We consider six client selection baselines: FedAvg \cite{mcmahan2017communication}, S-FedAvg \cite{nagalapatti2021game}, FedEMD \cite{huang2022tackling}, FedProx \cite{li2020federated}, GreedyFed\cite{singhal2024greedy}, and PoC \cite{cho2022towards}. FedAvg applies random sampling in each round. S-FedAvg and GreedyFed combine SV-based methods with client selection. FedProx and PoC propose mechanisms regarding data heterogeneity in FL. FedEMD combines EMD distance with client selection in the presence of \mave.

\begin{figure}[t]
  \centering
    \begin{subfigure}{0.238\textwidth}
    \centering
    \includegraphics[width=\linewidth]{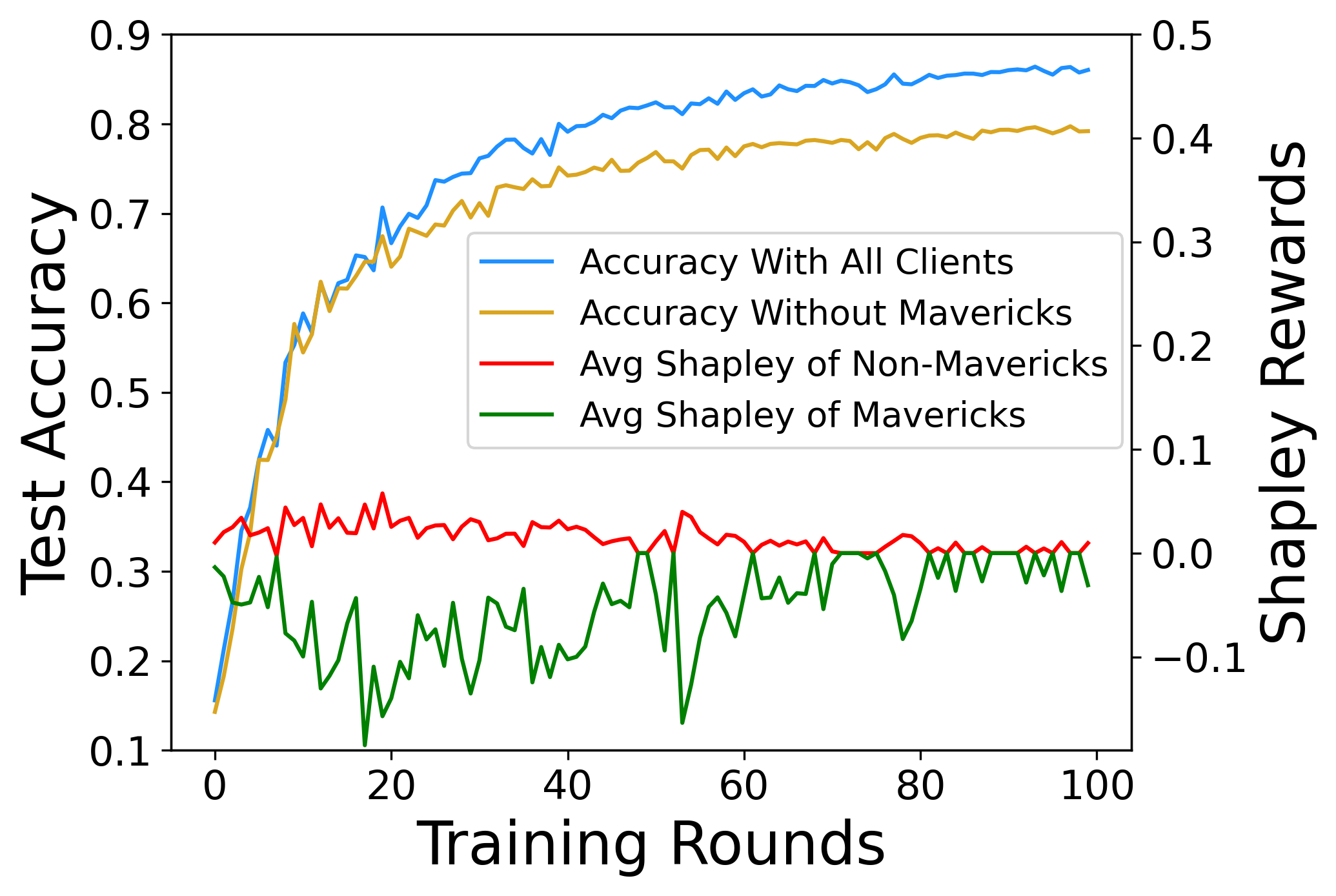}
    \caption{FedAvg (Original)}\label{fig:5_users_GTG_fedavg}
  \end{subfigure}
  \begin{subfigure}{0.238\textwidth}
    \centering
    \includegraphics[width=\linewidth]{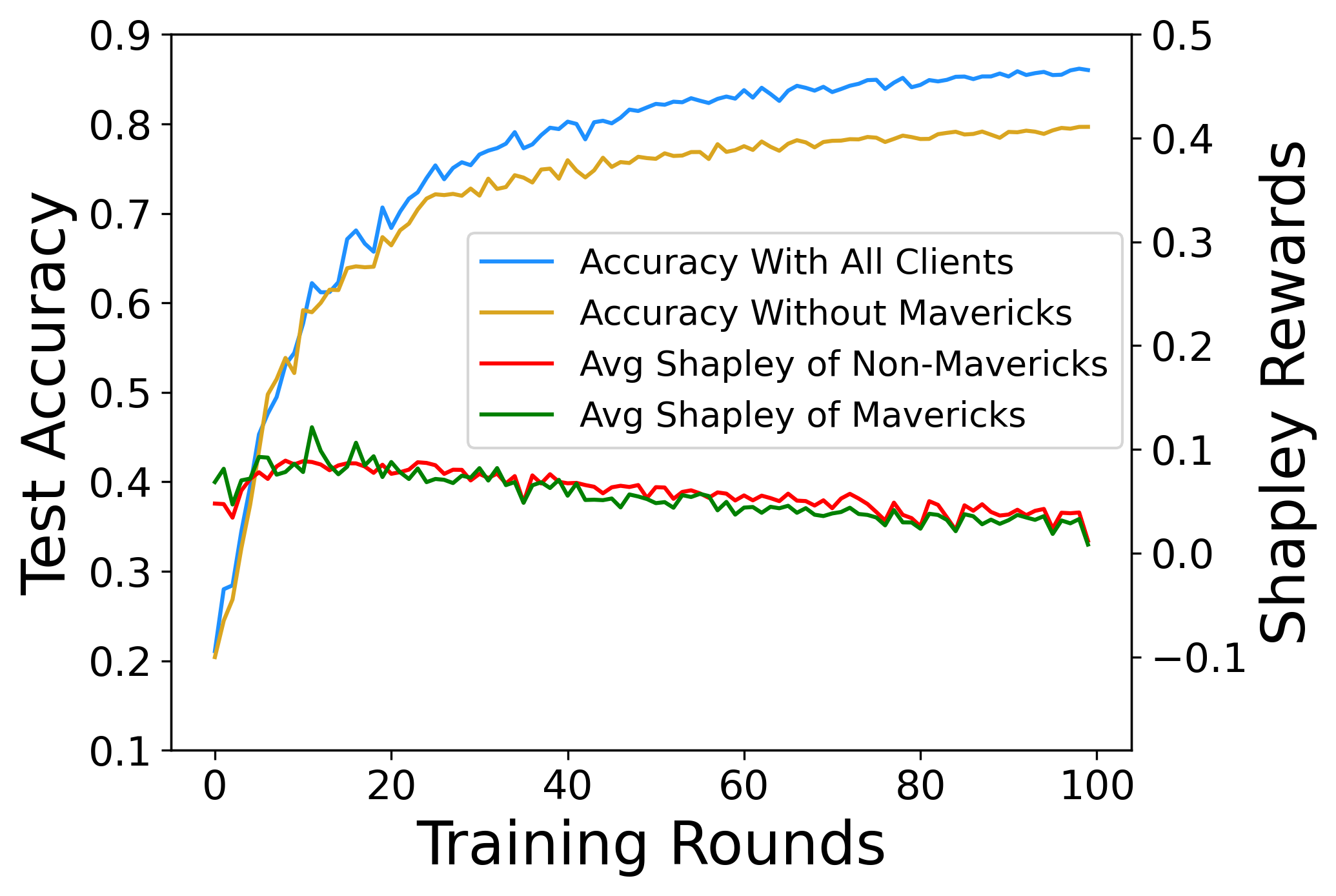}
    \caption{\algoname (Our method)}    \label{fig:5_users_GTG_fedccs}
  \end{subfigure}
  \vspace{-1mm}
  \caption{Comparison of test accuracy and Shapley rewards with 5 clients (w/o client selection) for the MNIST dataset using GTG-Shapley.}
 \label{fig:5_users_GTG}
  \vspace{-6mm}
\end{figure}

\begin{figure}[ht!]
    \centering
    \begin{subfigure}{0.238\textwidth}
        \centering
        \includegraphics[width=\linewidth]{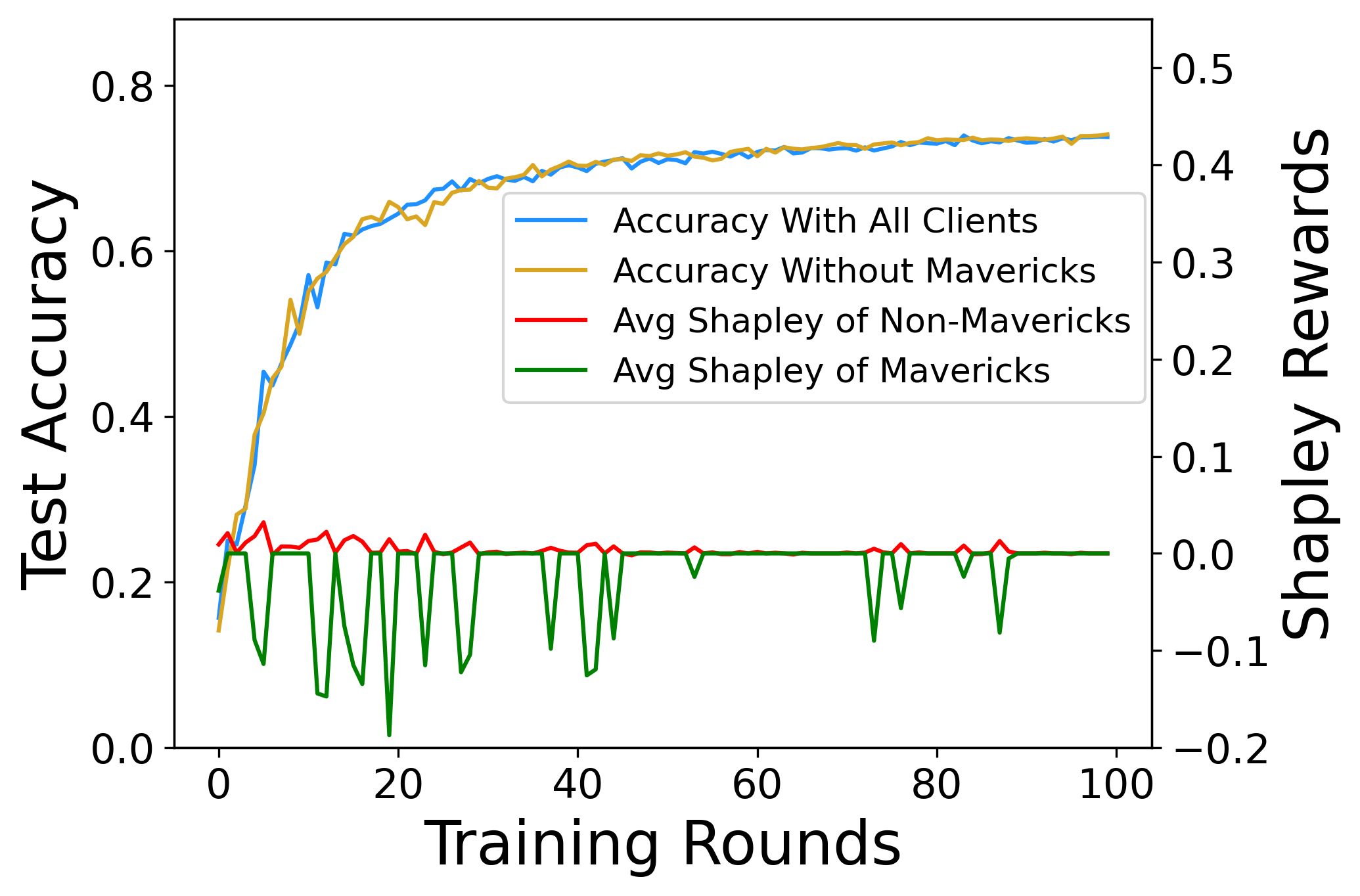}
\caption{FedAvg (Original)}
    \label{fig:GTG_MNIST_50_FedAVG}
    \end{subfigure}
    \begin{subfigure}{0.238\textwidth}
        \centering
        \includegraphics[width=\linewidth]{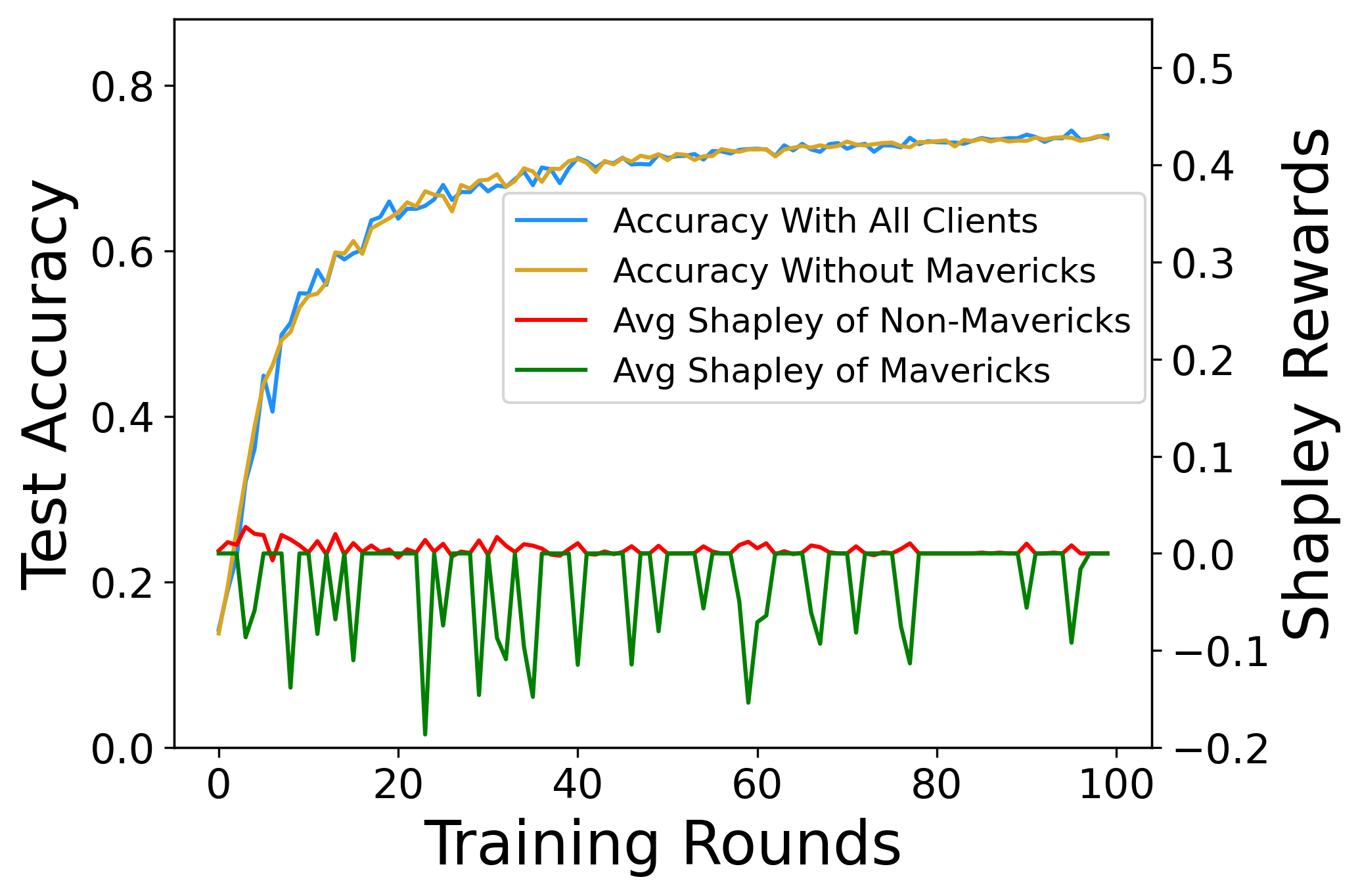}
\caption{S-FedAvg}
    \label{fig:GTG_MNIST_50_S_FedAVG}
    \end{subfigure}
    \begin{subfigure}{0.238\textwidth}
        \centering
        \includegraphics[width=\linewidth]{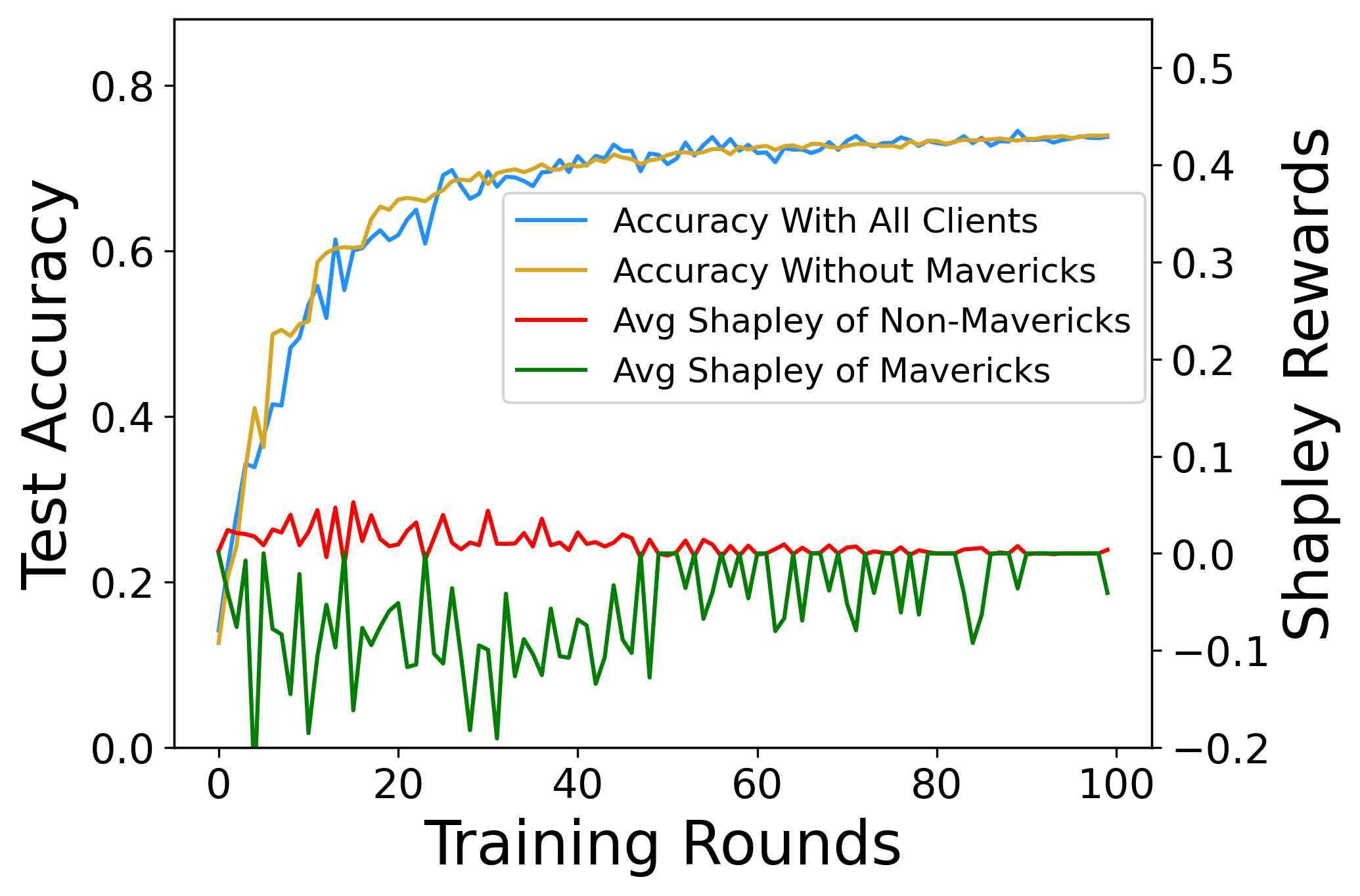}
\caption{FedEMD}
    \label{fig:GTG_MNIST_50_FedEMD}
    \end{subfigure}
\begin{subfigure}{0.238\textwidth}
        \centering
        \includegraphics[width=\linewidth]{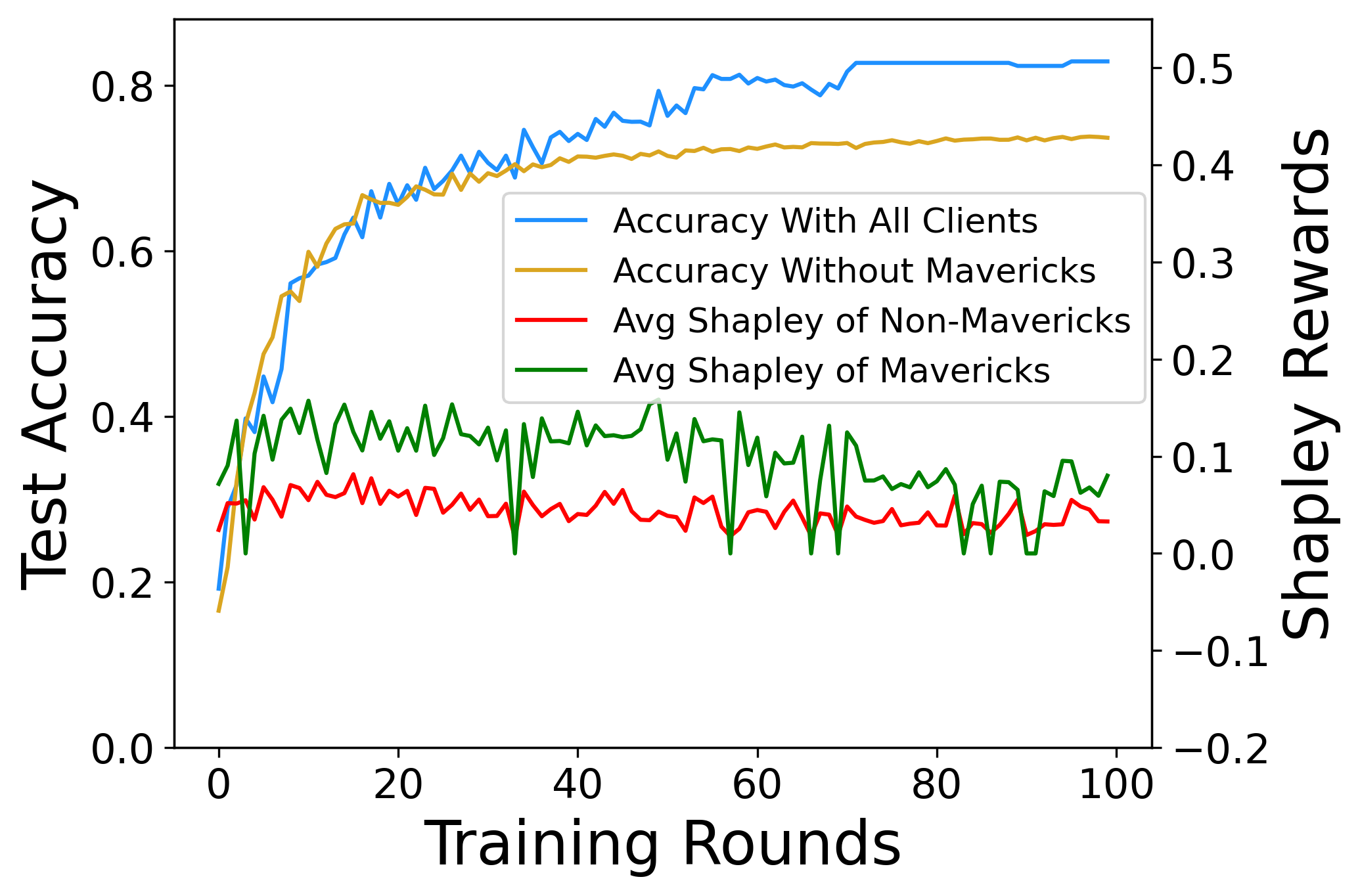}
\caption{\algoname (Our method)}
\label{fig:GTG_MNIST_50_FedCSV}
    \end{subfigure}
    \caption{Comparison of test accuracy and Shapley rewards with 50 clients (w/ client selection) for the MNIST dataset using GTG-Shapley for various client selection techniques.}
    \label{fig:GTG_MNIST_50}
    \vspace{-5mm}
\end{figure}

\noindent \textbf{Fairer Shapley (Reward) Distribution.}
Fig.~\ref{fig:5_users_GTG} illustrates how the test accuracy and SR change during training in the 5 client setting (w/o client selection). We see in Fig.~\ref{fig:5_users_GTG} that \mave helps increase the model accuracy. Despite this benefit of training with \mave, we observe that the average SR of \mave is considerably lower than that of non-\mave in FedAvg when the rewards are based on the SVs. In contrast, Fig.~\ref{fig:5_users_GTG_fedccs} exhibits a fairer SR for \mave.  In Fig.~\ref{fig:GTG_MNIST_50}, when considering the scenario with 50 clients (w/ client selection), \algoname assigns higher rewards to \mave than all baseline methods. %
In these experiments, we use the GTG-Shapley \cite{liu2022gtg} for SV computation.
We deduce that \algoname shows effectiveness for fairer SR distribution for \mave and non-\mave in both settings (w/ and w/o client selection), thanks to the class-wise SV-based rewards as in (\ref{Shapley_rewards}).

\noindent \textbf{Improved Model Performance.}
Figs.~\ref{fig:GTG_MNIST_50_FedAVG}, \ref{fig:GTG_MNIST_50_S_FedAVG} and \ref{fig:GTG_MNIST_50_FedEMD} display a similar test accuracy in the settings of {\em All Clients} and {\em Without \mave} for FedAvg, S-FedAvg, and FedEMD, respectively. On the other hand, in Fig.~\ref{fig:GTG_MNIST_50_FedCSV}, our method \algoname demonstrates an elevated accuracy in {\em All Clients} setting compared to the {\em Without \mave} setting, illustrating the effective utilization of \mave during FL training under the proposed approach.

\noindent \textbf{Comparisons with Baselines.}
Our proposed method, \algoname, outperforms the baselines in both SR and utility metrics. In regards to the SR, \algoname computes a fairer SR for all clients by considering class-wise SVs and class difficulties $\beta$. If rare classes owned by \mave perform poorly on the validation dataset, our mechanism increases the $\beta$ associated with these rare classes. Hence, our system boosts fairer SR for the \mave. Alternatively, in the baseline methods, only S-FedAvg and GreedyFed adopt SV in their client selection process but none of them considers the \mave settings. In those SV-based methods, the low SR %
of \mave decreases their selection probability during training, resulting in under-utilization of \mave. %
Since \algoname can effectively select the most contributing clients, it successfully selects \mave and shows an increased model accuracy, as shown in Fig.~\ref{fig:GTG_MNIST_50_FedCSV}. FedEMD applies a decreasing selection probability of \mave as iterations progress and our approach differs from FedEMD by not relying on the distance of local \& global data distributions. Instead, we prioritize the class-wise contribution of each client during the selection process, thus, our method achieves fairer SR and improved accuracy compared to FedEMD (see more comparisons in Appendix~\ref{app:additional_experiments}). %

\section{Conclusion and Future Directions}

The selection of clients plays a pivotal role in achieving success in FL, as it allows for the optimization of the utility derived from diverse model updates, specifically in the presence of Mavericks. In this work, we propose \algoname, a Maverick-aware Shapley valuation
mechanism for client selection in FL that not only fairly evaluates the contributions of the \mave but also effectively selects the most contributing clients in each training round. Our proposed \algoname achieves better model performance and fairer Shapley Rewards distribution compared to the existing methods.

\noindent \textbf{Future Directions:} \algoname does not consider potential attacks in the presence of Mavericks, particularly how attackers might exploit the system (e.g., attackers or outliers can pretend as Mavericks or the Mavericks themselves can take advantage of the system). Future research should focus on investigating potential attacks, such as poisoning and adversarial attacks, that target Maverick-friendly FL systems.

\clearpage
{\footnotesize \bibliographystyle{unsrt}
\bibliography{ref}}

\appendix
\label{appendix}

\subsection{Related Works}\label{app:related}
\noindent\textbf{Client Selection.} 
In an FL system, clients show different degrees of heterogeneity in data distribution and system resources. The vanilla mechanism FedAvg \cite{mcmahan2017communication} that randomly samples clients in each round may not fully leverage the diverse local updates from heterogeneous clients \cite{fu2023client}. Various selection methods have been proposed to deal with heterogeneity in FL to improve the model performance \cite{nagalapatti2021game, huang2022tackling, singhal2024greedy, cho2022towards}. S-FedAvg \cite{nagalapatti2021game} combines FedAvg with SV and empowers the server to select relevant clients with high probability. GreedyFed \cite{singhal2024greedy} greedily selects the most contributing clients in each round by employing a fast Shapley approximation algorithm named the GTG-Shapley \cite{liu2022gtg}. In Power-of-Choice (POC) \cite{cho2022towards}, %
authors propose a client scheduling strategy that selects the client models with the highest loss in each round. Common to all these methods is the fact that none of them considers the \mave. Recently, authors in \cite{huang2022tackling} introduced the concept of \mave and proposed FedEMD to adaptively select clients based on the Wasserstein distance between the local and global data distributions. Although FedEMD increases the probability of selecting the \mave, it does not provide a solution to fairly evaluate the contribution of the \mave.

\noindent\textbf{Contribution Evaluation via Gradient Shapley Methods.} 
SV-based methods are widely employed in FL to compute the contributions of the participating clients \cite{wei2020efficient, song2019profit, liu2022gtg}. Despite its prominence in the game theory literature, in the context of ML, SV \cite{shapley1971cores} is not practical as it requires retraining from scratch considering each client permutation. Gradient Shapley methods aim to eliminate the lengthy retraining of FL models by utilizing gradient updates of the clients to approximate the FL sub-models for various clients permutations in the SV computation. Reference \cite{song2019profit} proposes two gradient Shapley methods: one-round (OR) and multi-round (MR). OR calculates the SV once after the training while MR calculates the SV in every FL round. Truncated Multi-Rounds Construction (TMR) \cite{wei2020efficient} eliminates unnecessary FL sub-model reconstructions by adding a decay factor. In Guided Truncation Gradient Shapley (GTG-Shapley) \cite{liu2022gtg}, authors design a guided Monte Carlo sampling approach combined with truncation techniques to further improve the computation efficiency. Despite these efforts to efficiently and accurately approximate the SV, previous works \cite{huang2022tackling, buyukates2023proof} showed that the current SV-based methods are unable to fairly assess the contributions of the \mave. Motivated by these, one of our goals in this work is to propose a SV-based contribution score that can appreciate the contributions of both Maverick and non-Maverick clients. We then use the accumulated contribution scores of the clients to perform intelligent client selection in each round to better utilize the \mave during training and improve the model performance.

\subsection{Maverick GTG-Shapley}\label{app:maverick-getg}

The employed class-wise Shapley value computation technique in \algoname, i.e., Maverick-Shapley, is compatible with the existing SV approximation approaches. In Algorithm~\ref{alg:CGTG-Shapley}, we describe the class-wise Shapley computation by using the GTG-Shapley \cite{liu2022gtg} technique.

\begin{algorithm}[h]
\caption{Maverick GTG-Shapley}
\label{alg:CGTG-Shapley}
\small
\DontPrintSemicolon
\SetAlgoNoEnd
\SetAlgoNoLine
\SetKwFunction{ModelAverage}{ModelAverage}

\textbf{Input}: Updated client models $\{\bm{w}_i^{t}\}_{i \in \mathcal{K}^t}$; current server model $\bm{w}^{t}$;  validation dataset at server $D_{val}$; class-wise accuracy function $\mathcal{V}_{class}(\cdot)$; $\mathcal{M}$: set of class labels.\\
 \textbf{Hyperparameters:} Error threshold $\epsilon_b$, $\epsilon_i$, temperature T. \\
 \textbf{Initialize:} $\phi_i=0, \forall i\in \mathcal{K}^t, r=0$\\
 Compute $w^{t+1} = \ModelAverage(n_i,\, \{\bm{w}_i^{t}\}_{i \in \mathcal{K}^t})$\\
 $v_0 = \mathcal{V}_{class}(w^{t};\mathcal{D}_{\textrm{val}}),\, v_N = \mathcal{V}_{class}(w^{t+1};\mathcal{D}_{\textrm{val}}),\, $\\
 \textit{\# between round truncation }\\
 \textbf{if} {$|v_N-v_0|>\epsilon_b$} \textbf{then} \\
 \SP \textbf{while} {Convergence criteria not met} \textbf{do} \\
\SP \SP  $r=r+1$\\
\SP \SP \For{client $i\in \mathcal{K}^t$ }{
 \SP permute $\mathcal{K}^t \setminus \{i\}:\pi^r[0] = i,\pi^r[1:n] $\\
 \SP $v_0^r=v_0$\\
\SP \textit{\# within-round truncation }\\
 \SP \For{$j=1,\dots,n$}{
 \textbf{if} $|v_N-v^r_{j-1}| \ge \epsilon_i$\\
 \SP \SP  $H=\pi^r[:j]$\\
 \SP \SP  $\widetilde{w}_H=\ModelAverage(\{\bm{w}_i^{t}\}_{i \in H},\bm{w}^{t})$\\
 \SP \SP $v^r_j=\mathcal{V}_{class}(\widetilde{w}_H;\mathcal{D}_{\textrm{val}})$\\
 \textbf{else} \\
 \SP \SP $v^r_j=v^r_{j-1}$\\
 \SP \SP \For{class $c$ $\in$ $\mathcal{M}$}{
 \SP \SP  $\phi_{\pi^r[j]}^c = \frac{r-1}{r}\phi_{\pi^r[j]}^c  + \frac{(v_j^{r,c} -v_{j-1}^{r,c})}{r}$}}}

\textit{\# Find best clients set $\hat{\mathcal{K}}^{t}$ and its class-wise accuracy $\hat{v}$}\\
$\hat{\mathcal{K}}^{t}, \hat{v} \leftarrow \operatorname*{argmax}_H \sum \limits_{c \in \mathcal{M}} \mathcal{V}_{class}^c (\widetilde{w}_H;\mathcal{D}_{\textrm{val}})$ \\
\textit{\# Obtain  class difficulty $\beta$}\\
$\beta^c = \frac{exp(\frac{1 - \hat{v}^c}{T})}{\sum \limits_{c \in M}exp(\frac{1 - \hat{v}^c}{T})}, \forall c \in M$ \\

\Return{$\phi, \beta, \hat{\mathcal{K}}^{t}$}
\end{algorithm}

\subsection{\algoname Parameters \& Notation} \label{app:params}

Table~\ref{tab:params_table} lists the parameters and notation we use in \algoname.

\begin{table}[t!]
\centering
\smallskip\noindent
\begin{tabular}{@{}cl@{}}
\toprule
\centering
\textbf{Notation} & \textbf{Description}                                \\ \midrule
$(\bm{x}, y)$               & $\bm{x}$ is the feature vector and $y$ is the corresponding label                       \\
$\mathcal{K}$                & Set of clients such that $\mathcal{K} = \{1,2,...,I\}$ \\
$\mathcal{M}$             & Set of class labels $\mathcal{M} = \{1,2,...,C\}$ \\
$\eta_i$             & Learning rate at client $i$   \\  
$\mathcal{D}_i$    & Local dataset of client $i$  \\
$\mathcal{D}_{val}$  & Validation dataset at the server\\
${\bm{w}}$      & Learnable weights of the global model\\
${\bm{w}}_i$        & Local model of client $i$ ${\bm{w}}_i$\\
$\mathcal{V}_{class}(\cdot)$ & Class-wise accuracy function\\
$\phi$           & Class-wise Shapley value vector (including all clients)\\
$\phi_i$        & Class-wise Shapley value of the $i$-th client \\
$\beta$            & Class difficulty vector (including all classes)\\
$\beta^c$            & Class difficulty of class $c$\\
$S_{i}^c$          & Accumulated Shapley value of client $i$ for class $c$ \\
$\alpha$            & Decay factor for the accumulated Shapley value $S_{i}^c$ \\
$\hat{S}_{i}^c$          & Contribution score of client $i$\\
$P_{\hat{S}}$          & Selection probability vector for client selection (including all clients)\\
$P_{\hat{S},i}$          & Selection probability of client $i$ for client selection

\\
\hline
\\
$T$          & Number of FL training rounds \\
$t$               & Index of FL round, $t=0,1,2,...T-1$ \\
$E$                 & Number of local epochs \\
$B$                & Mini-batch size\\
${\bm{w}^t}$      & Learnable weights of the global model in round $t$\\
${\bm{w}}_i^t$        & Local model of client $i$ in round $t$\\
$\mathcal{K}^t$      & Set of selected clients in round $t$ with selection strategy $\pi$ \\
$\hat{\mathcal{K}}^{t}$ & Best clients set with the highest class-accuracy in round $t$ \\
$n_i^t$             & Dataset size of the $i$-th client in round $t$\\

\bottomrule
\end{tabular}
\caption{\small Main parameters and notation.}

\label{tab:params_table}
\end{table}

\subsection{Additional Experiments} \label{app:additional_experiments}

\begin{figure*}[t!]
    \centering
    \begin{subfigure}{0.245\textwidth}
        \centering
        \includegraphics[width=\linewidth]{images/50_user/50_user_GTG_FedCSV.png}
\caption{FedMS (Our method)}
\label{fig:GTG_FedMS}
    \end{subfigure}
    \hfill
    \begin{subfigure}{0.245\textwidth}
        \centering
        \includegraphics[width=\linewidth]{{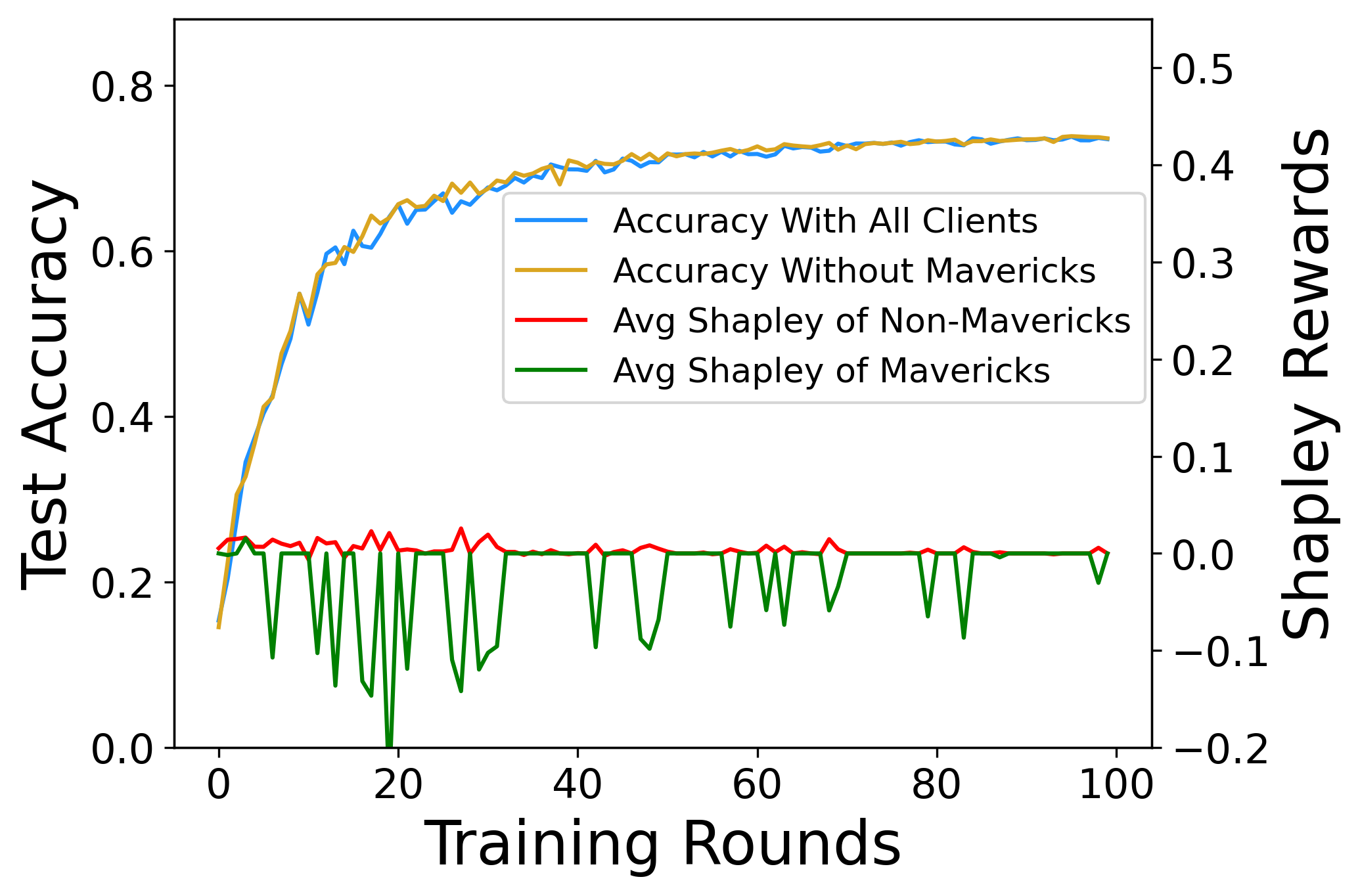}}
\caption{FedProx}
\label{fig:GTG_FedProx}
    \end{subfigure}
\begin{subfigure}{0.245\textwidth}
        \centering
        \includegraphics[width=\linewidth]{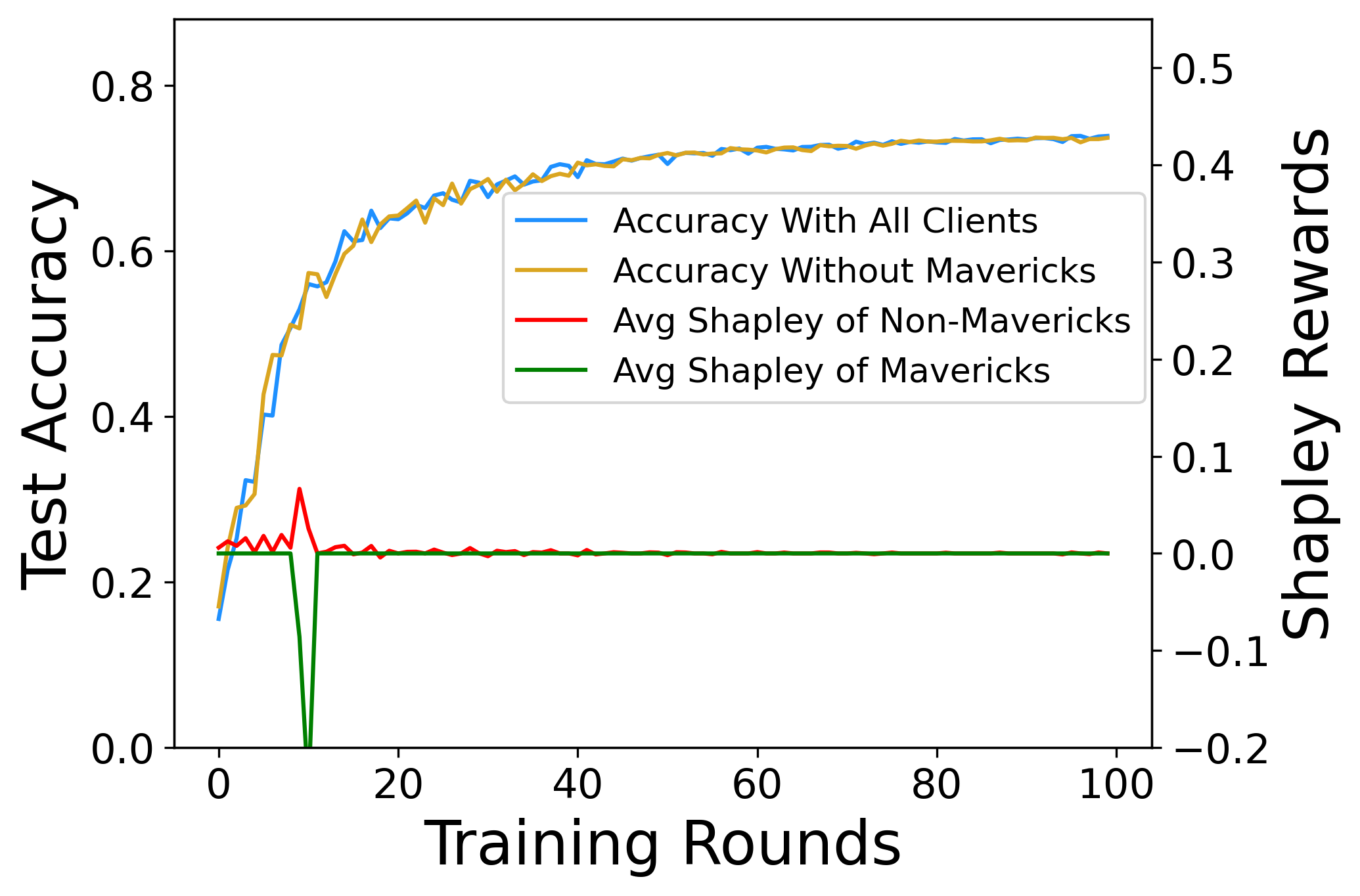}
\caption{GreedyFed}
\label{fig:GTG_Greedy}
    \end{subfigure}
    \hfill
\begin{subfigure}{0.245\textwidth}
        \centering
        \includegraphics[width=\linewidth]{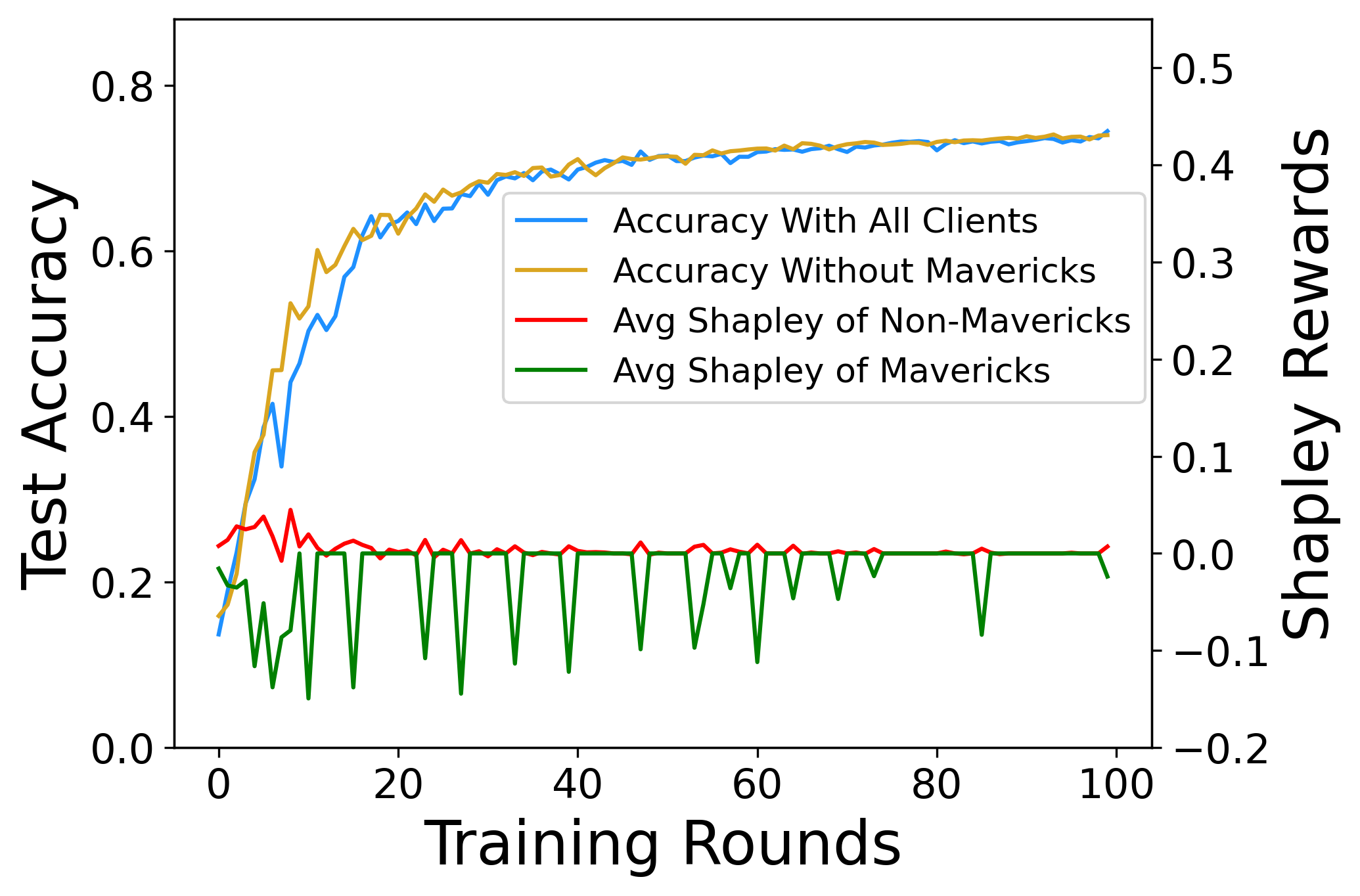}
\caption{PoC}
\label{fig:GTG_PoC}
    \end{subfigure}
    \hfill
    \caption{Comparison of test accuracy and Shapley rewards with 50 clients (w/ client selection) for the MNIST dataset using GTG-Shapley for various client selection techniques.}
    \label{fig:GTG_MNIST_appendix}
\end{figure*}

\begin{figure*}[t!]
    \centering
    \begin{subfigure}{0.245\textwidth}
        \centering
        \includegraphics[width=\linewidth]{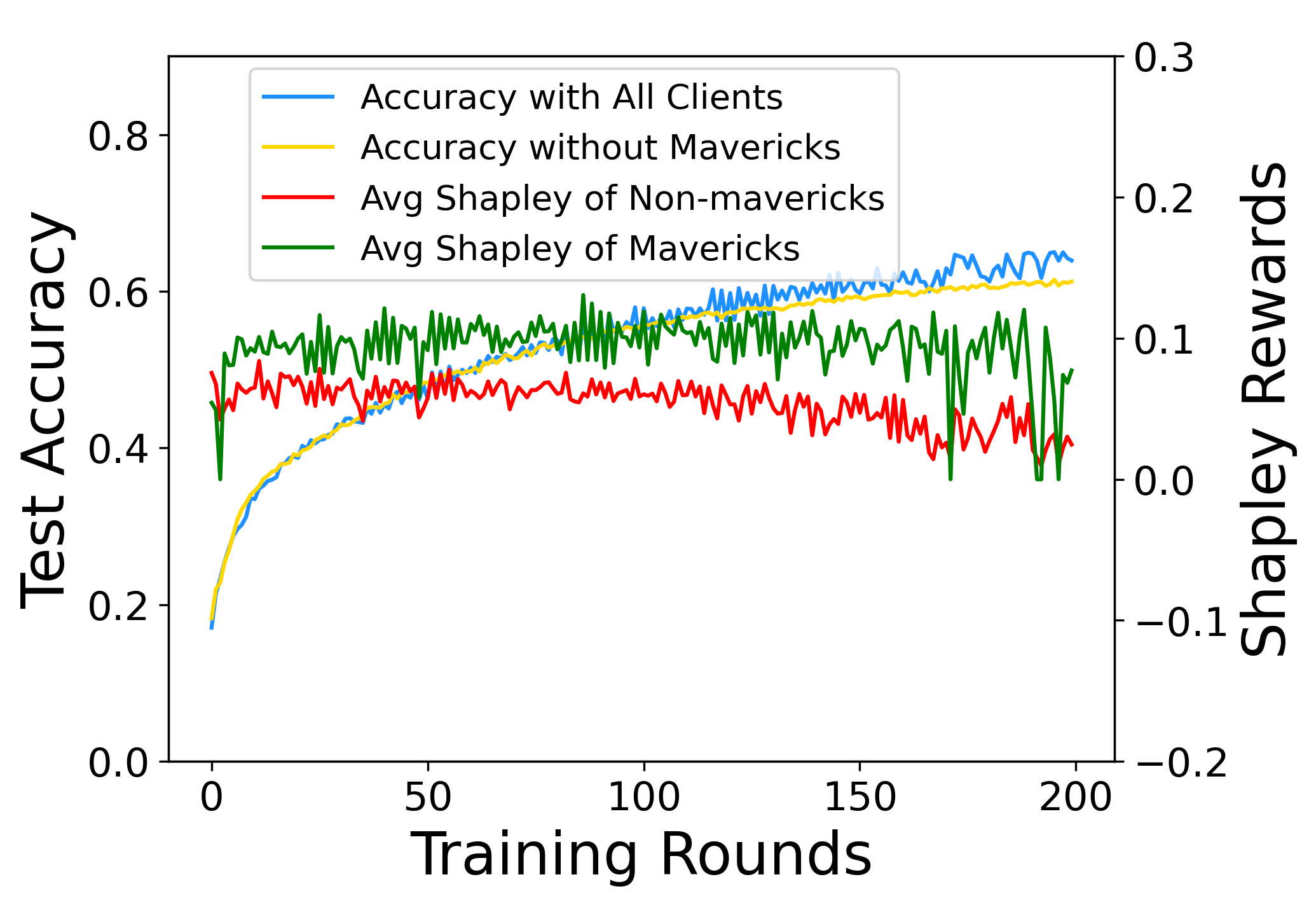}
\caption{FedMS (Our method)}
\label{fig:GTG_FedMS}
    \end{subfigure}
    \hfill
    \begin{subfigure}{0.245\textwidth}
        \centering
        \includegraphics[width=\linewidth]{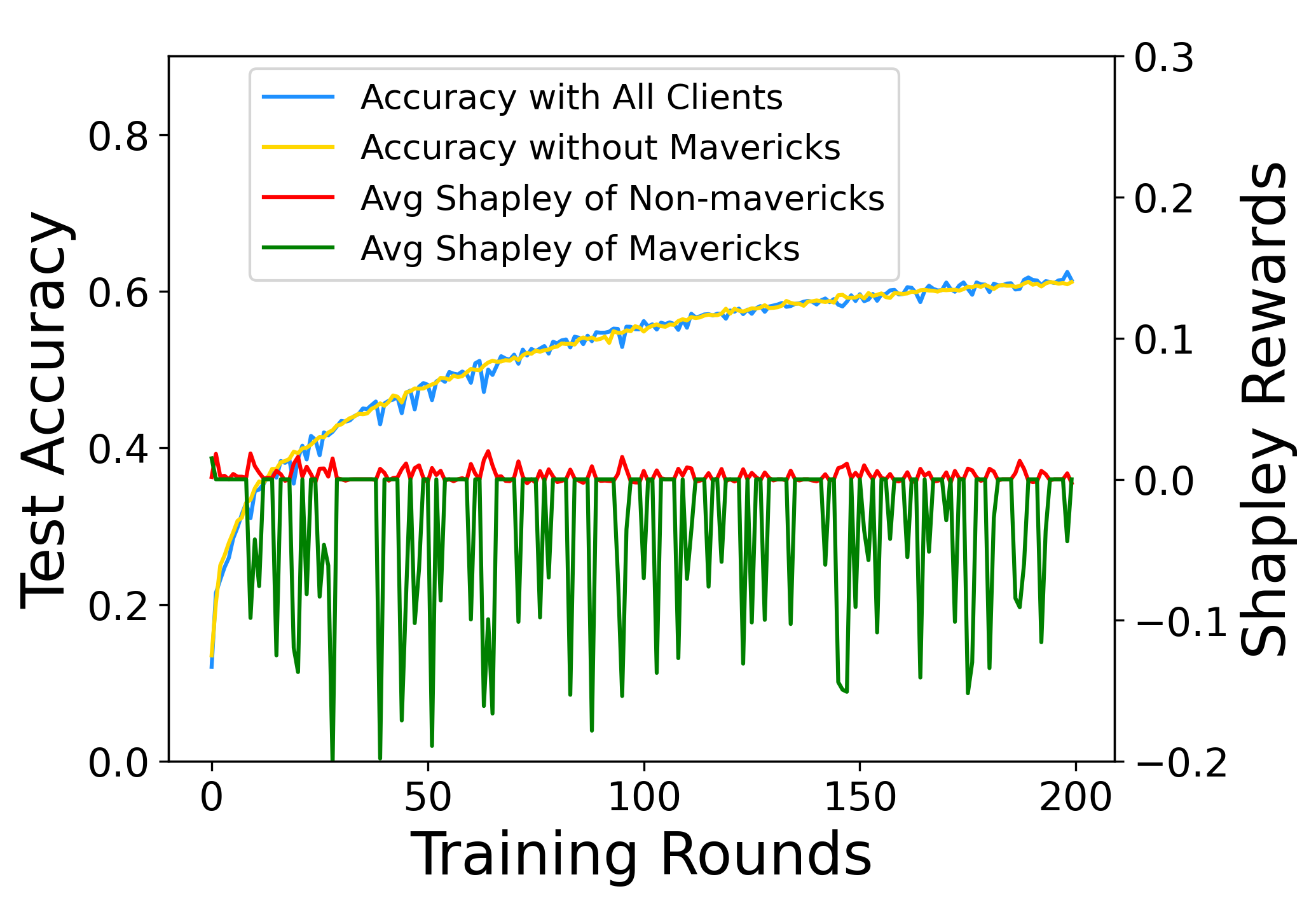}
\caption{FedAvg (original)}
\label{fig:GTG_FedAvg}
    \end{subfigure}
    \hfill
    \begin{subfigure}{0.245\textwidth}
        \centering
        \includegraphics[width=\linewidth]{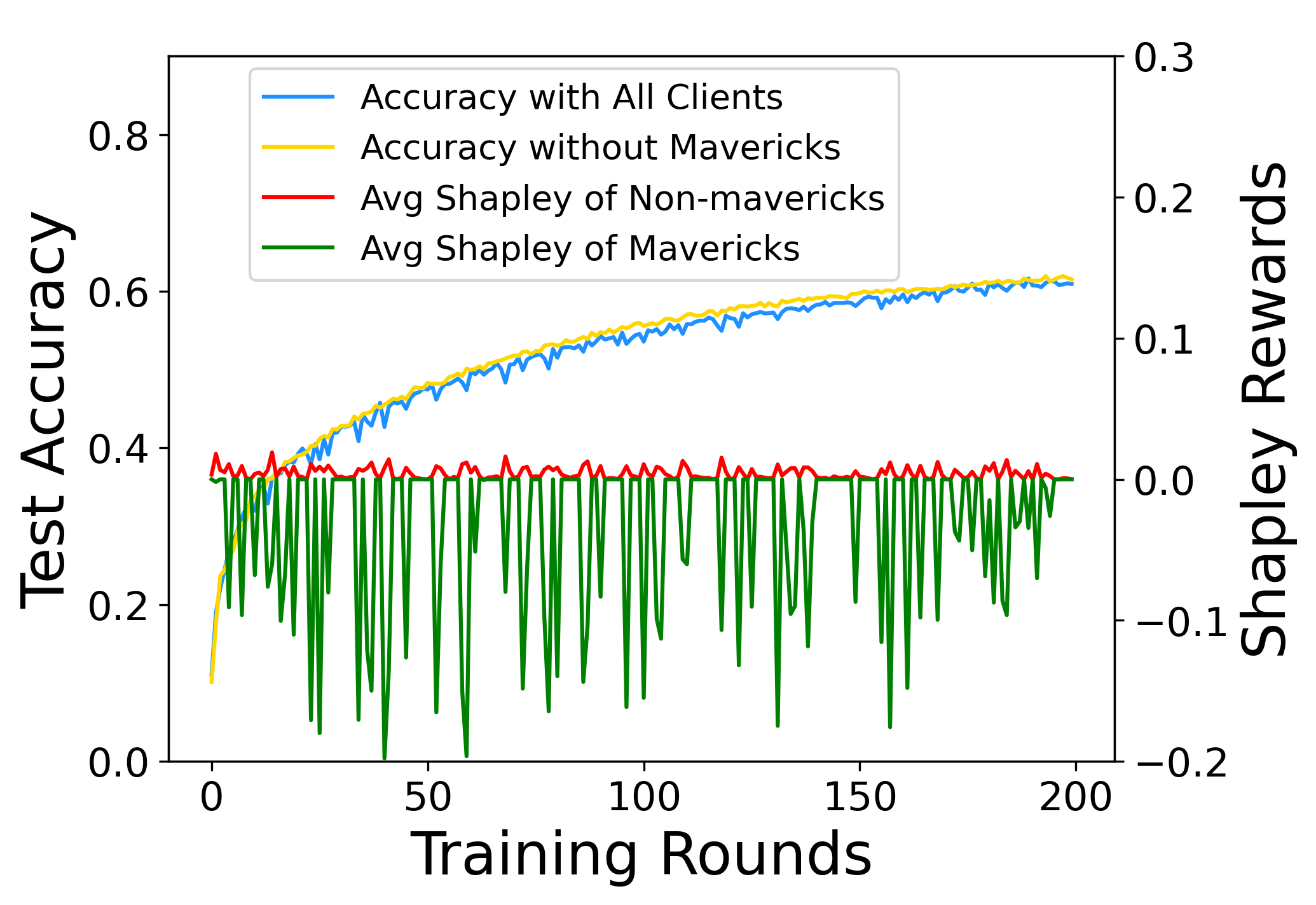}
\caption{S-FedAvg}
\label{fig:GTG_S_FedAvg}
    \end{subfigure}
    \hfill
    \begin{subfigure}{0.245\textwidth}
        \centering
        \includegraphics[width=\linewidth]{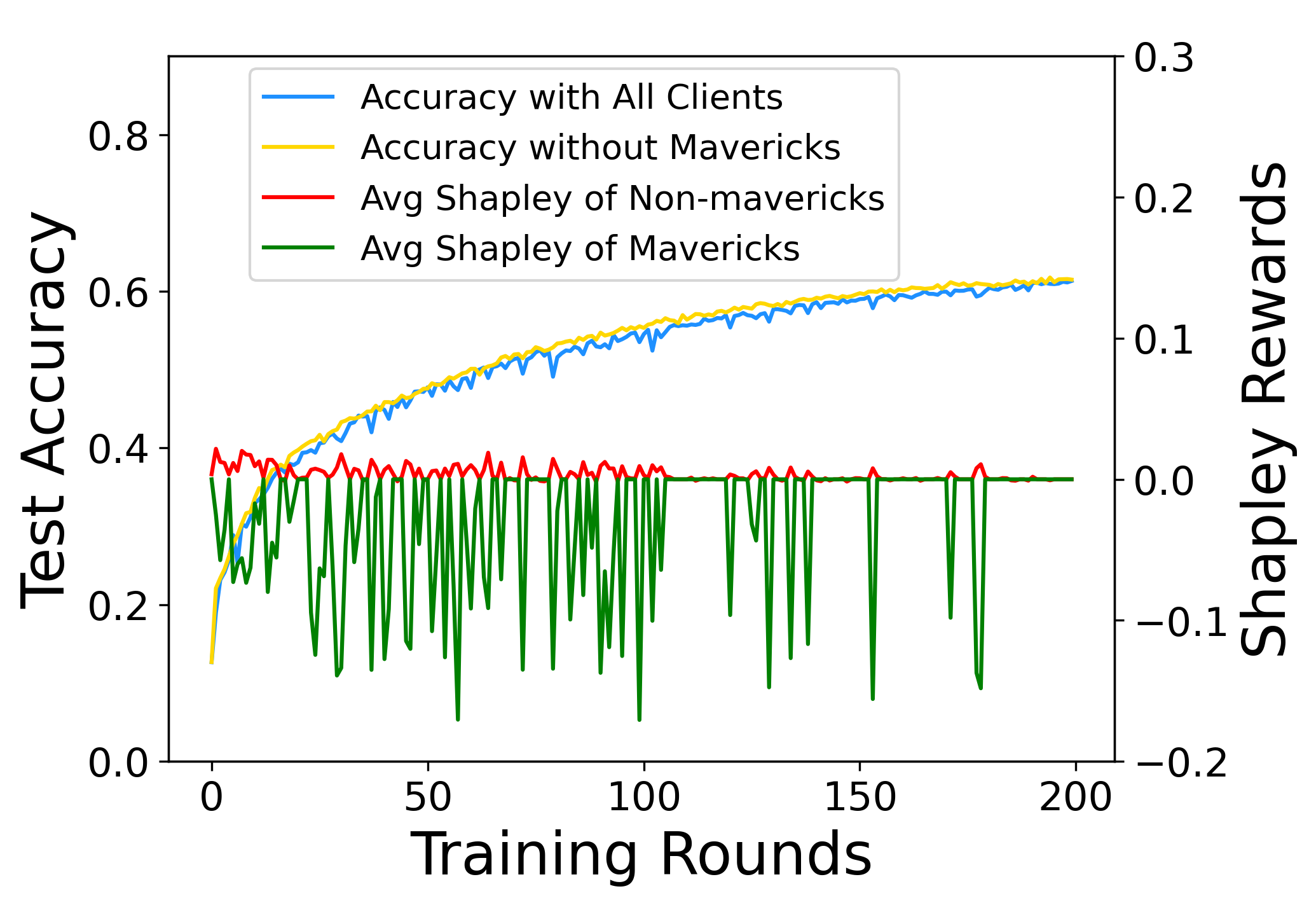}
\caption{FedEMD}
\label{fig:GTG_FedEMD}
    \end{subfigure}
    \centering
    \hfill
    \caption{Comparison of test accuracy and Shapley rewards with 50 clients (w/ client selection) for the CIFAR-10 dataset using GTG-Shapley for various client selection techniques.}
    \label{fig:GTG_Cifar10}
\end{figure*}

\begin{figure*}[t!]
    \centering
    \begin{subfigure}{0.245\textwidth}
        \centering
        \includegraphics[width=\linewidth]{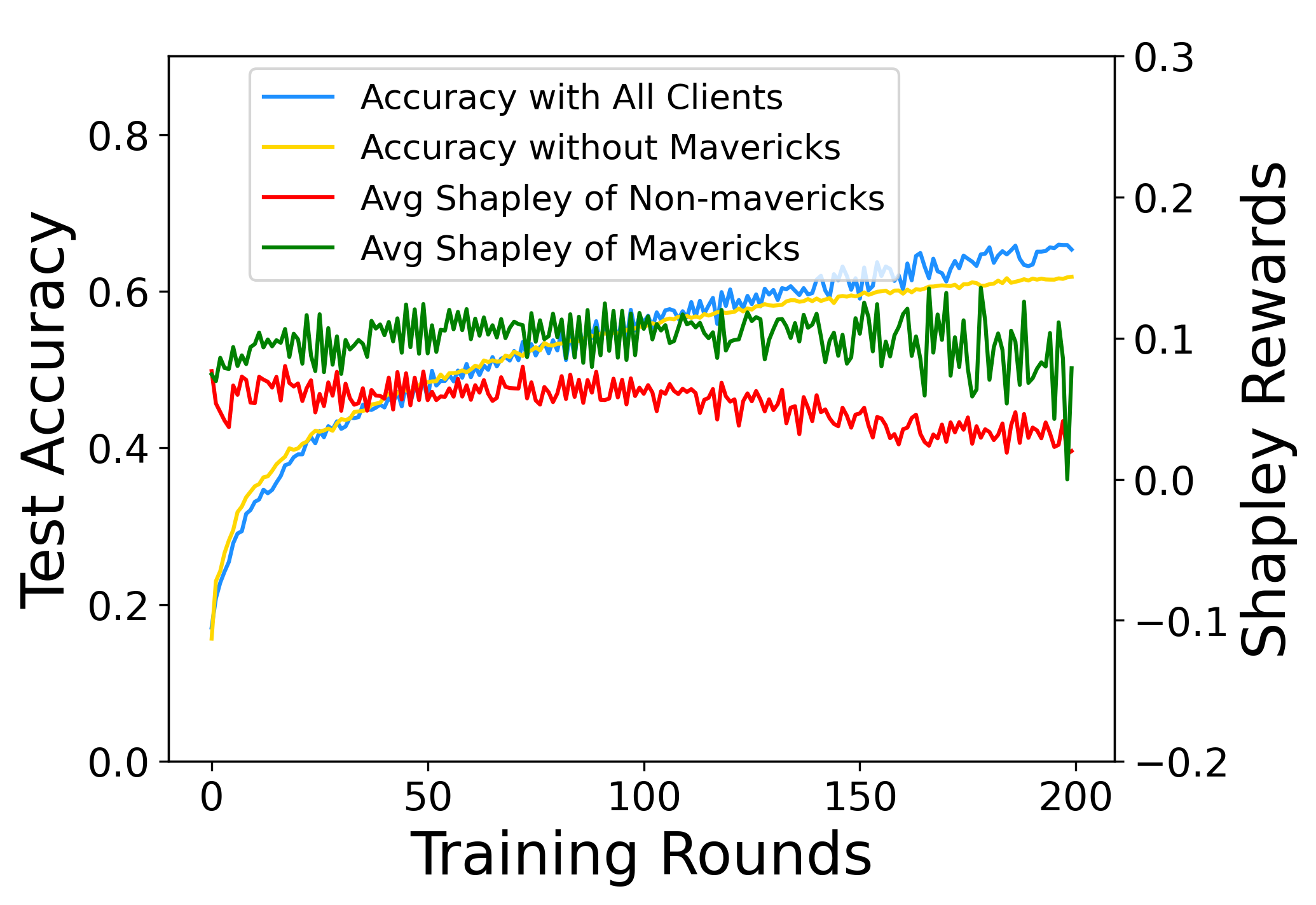}
\caption{FedMS (Our method)}
\label{fig:MR_FedMS}
    \end{subfigure}
    \hfill
    \begin{subfigure}{0.245\textwidth}
        \centering
        \includegraphics[width=\linewidth]{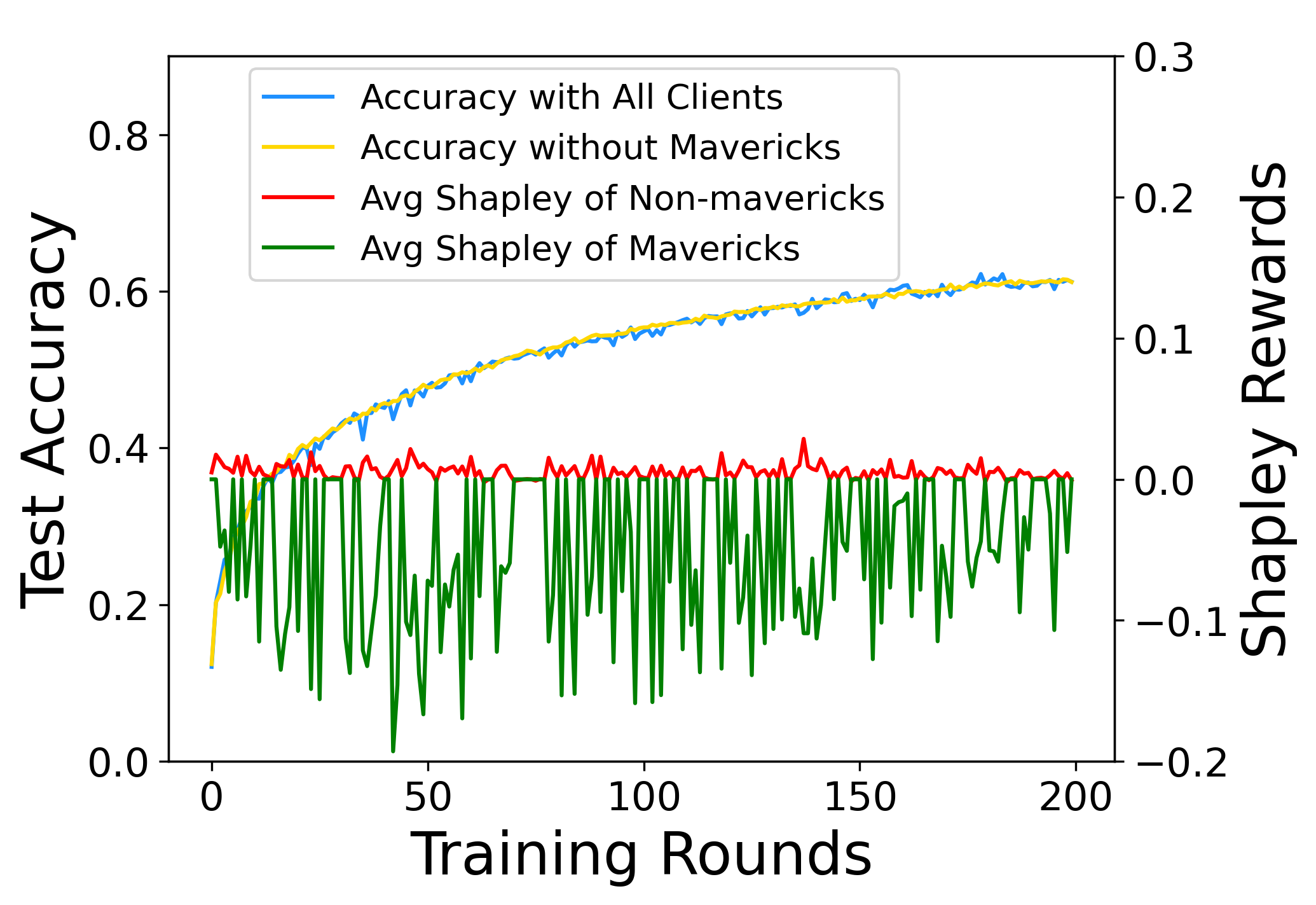}
\caption{FedAvg (original)}
\label{fig:MR_FedAvg}
    \end{subfigure}
    \hfill
    \begin{subfigure}{0.245\textwidth}
        \centering
        \includegraphics[width=\linewidth]{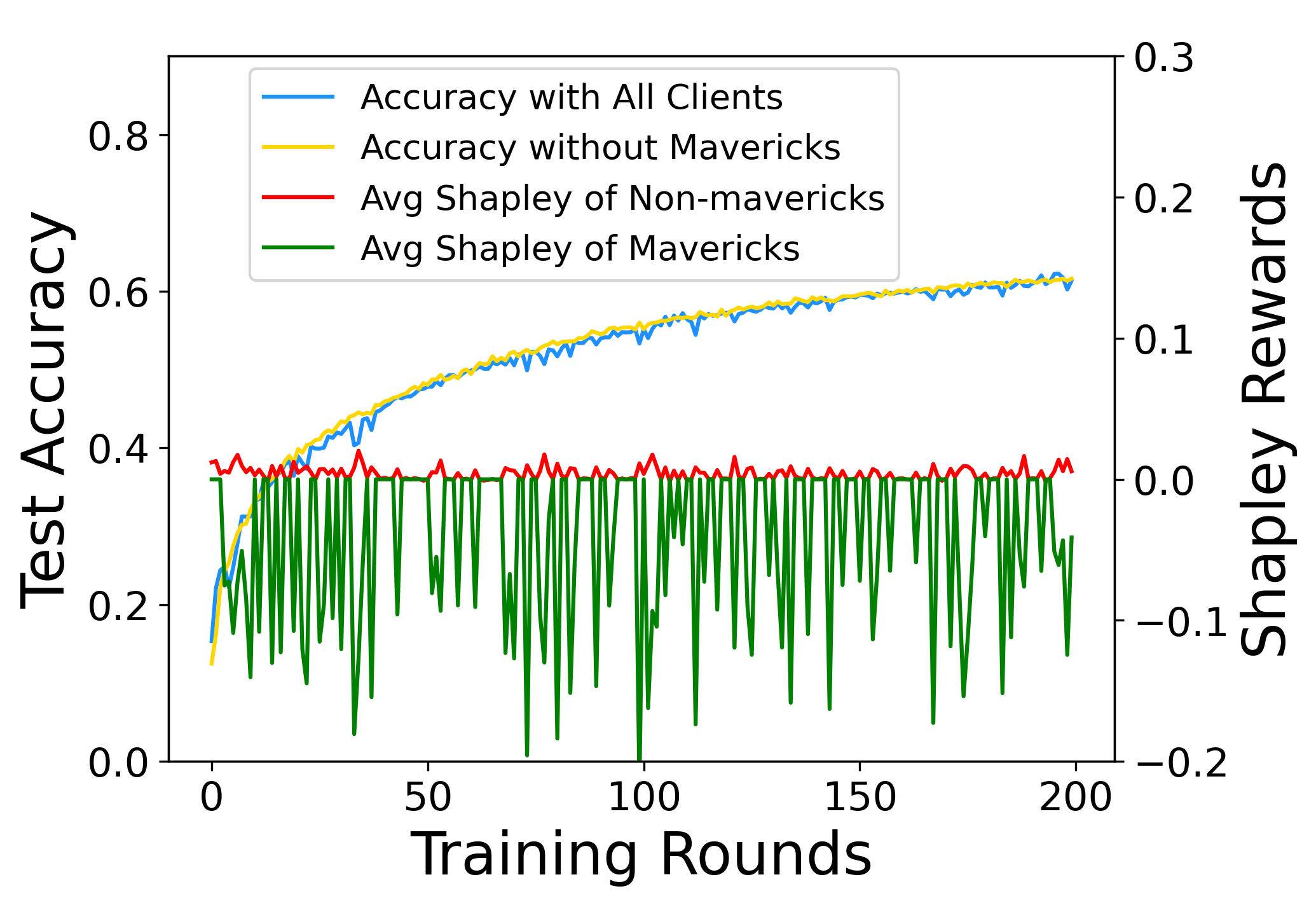}
\caption{S-FedAvg}
\label{fig:MR_S_FedAvg}
    \end{subfigure}
    \hfill
    \begin{subfigure}{0.245\textwidth}
        \centering
        \includegraphics[width=\linewidth]{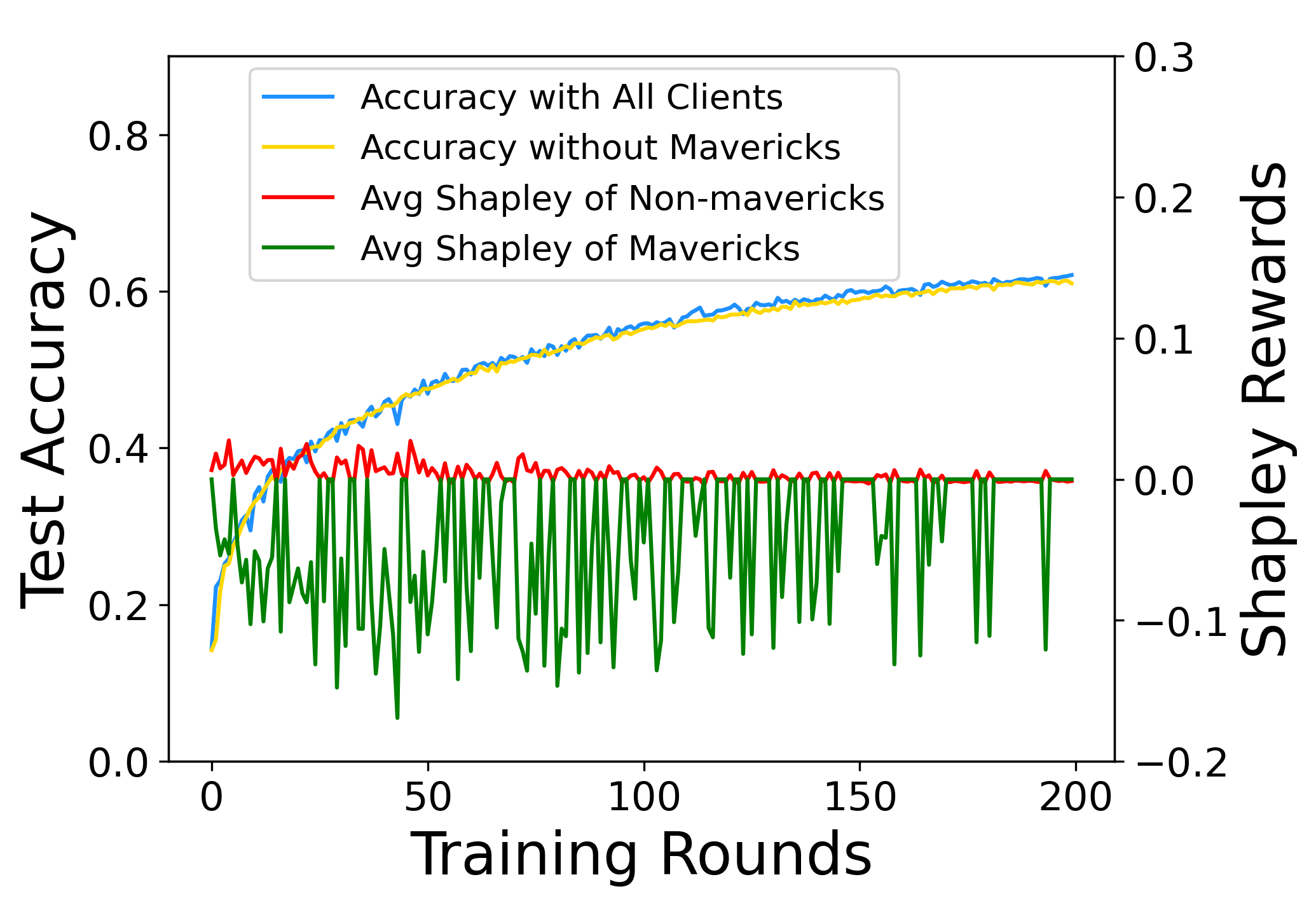}
\caption{FedEMD}
\label{fig:MR_FedEMD}
    \end{subfigure}
    \centering
    \hfill
    \caption{Comparison of test accuracy and Shapley rewards with 50 clients (w/ client selection) for the CIFAR-10 dataset using MR Shapley for various client selection techniques.}
    \label{fig:MR_Cifar10}
\end{figure*}

\begin{figure*}[t!]
    \centering
    \begin{subfigure}{0.245\textwidth}
        \centering
        \includegraphics[width=\linewidth]{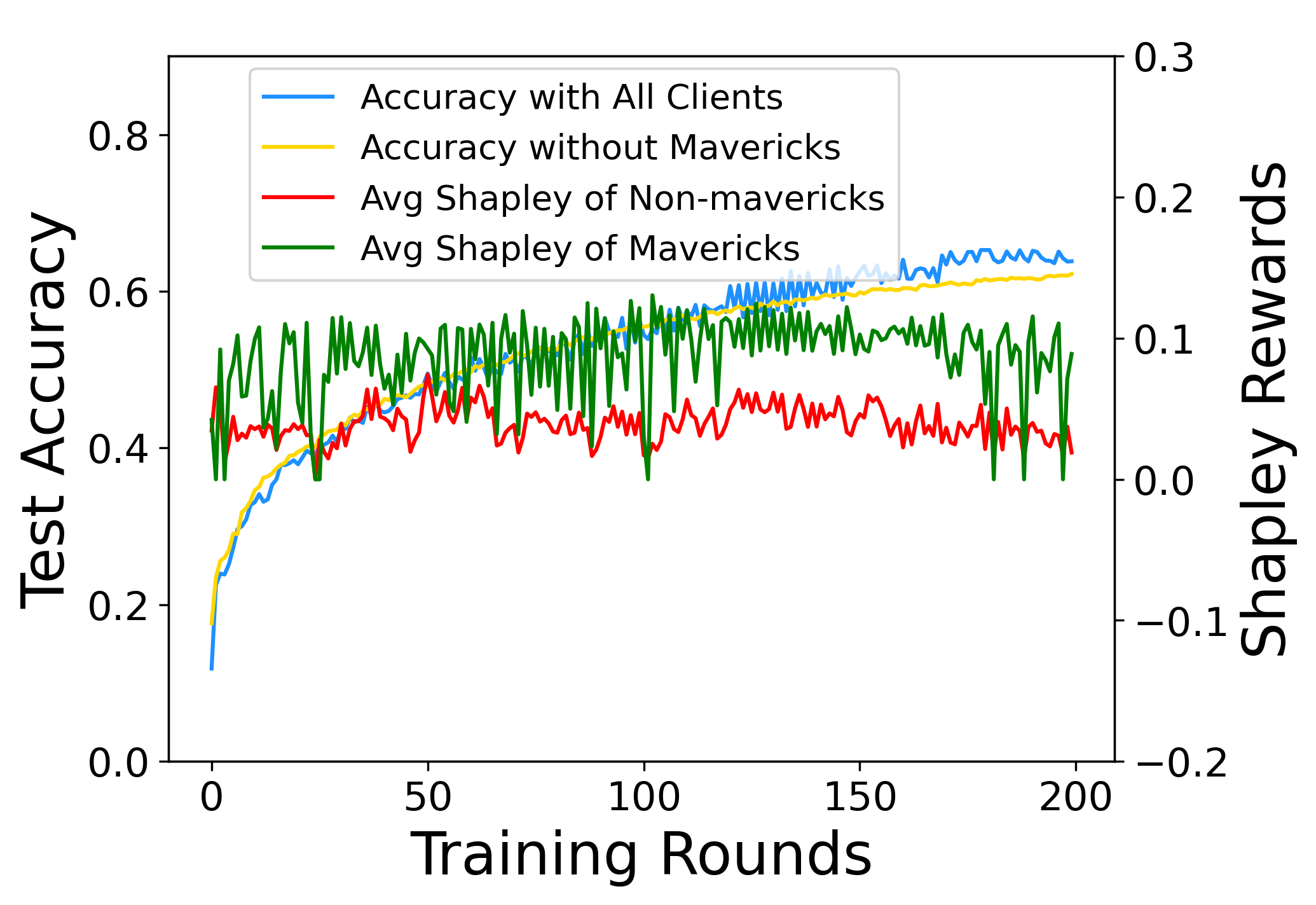}
\caption{FedMS (Our method)}
\label{fig:TMR_FedMS}
    \end{subfigure}
    \hfill
    \begin{subfigure}{0.245\textwidth}
        \centering
        \includegraphics[width=\linewidth]{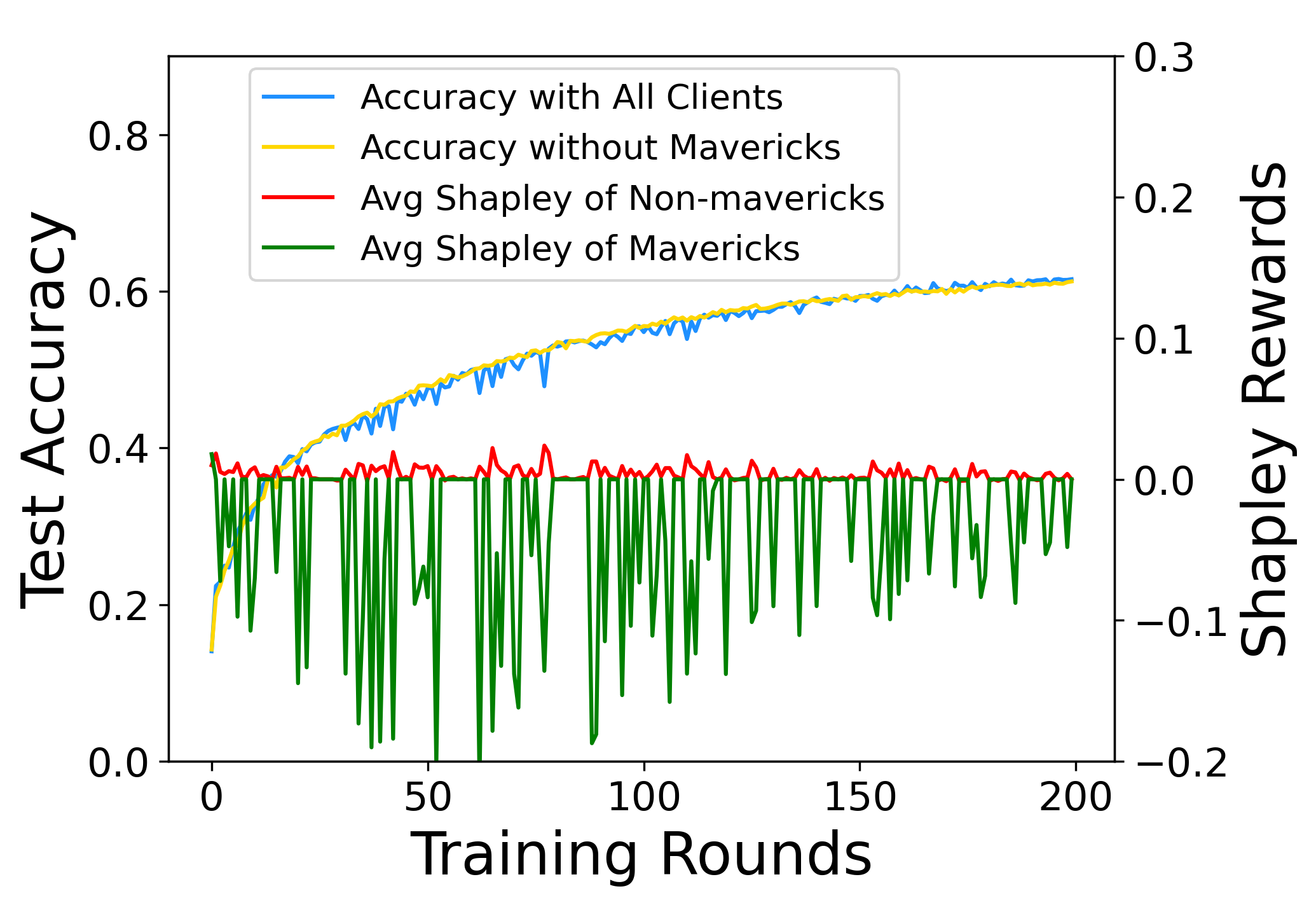}
\caption{FedAvg (original)}
\label{fig:TMR_FedAvg}
    \end{subfigure}
    \hfill
    \begin{subfigure}{0.245\textwidth}
        \centering
        \includegraphics[width=\linewidth]{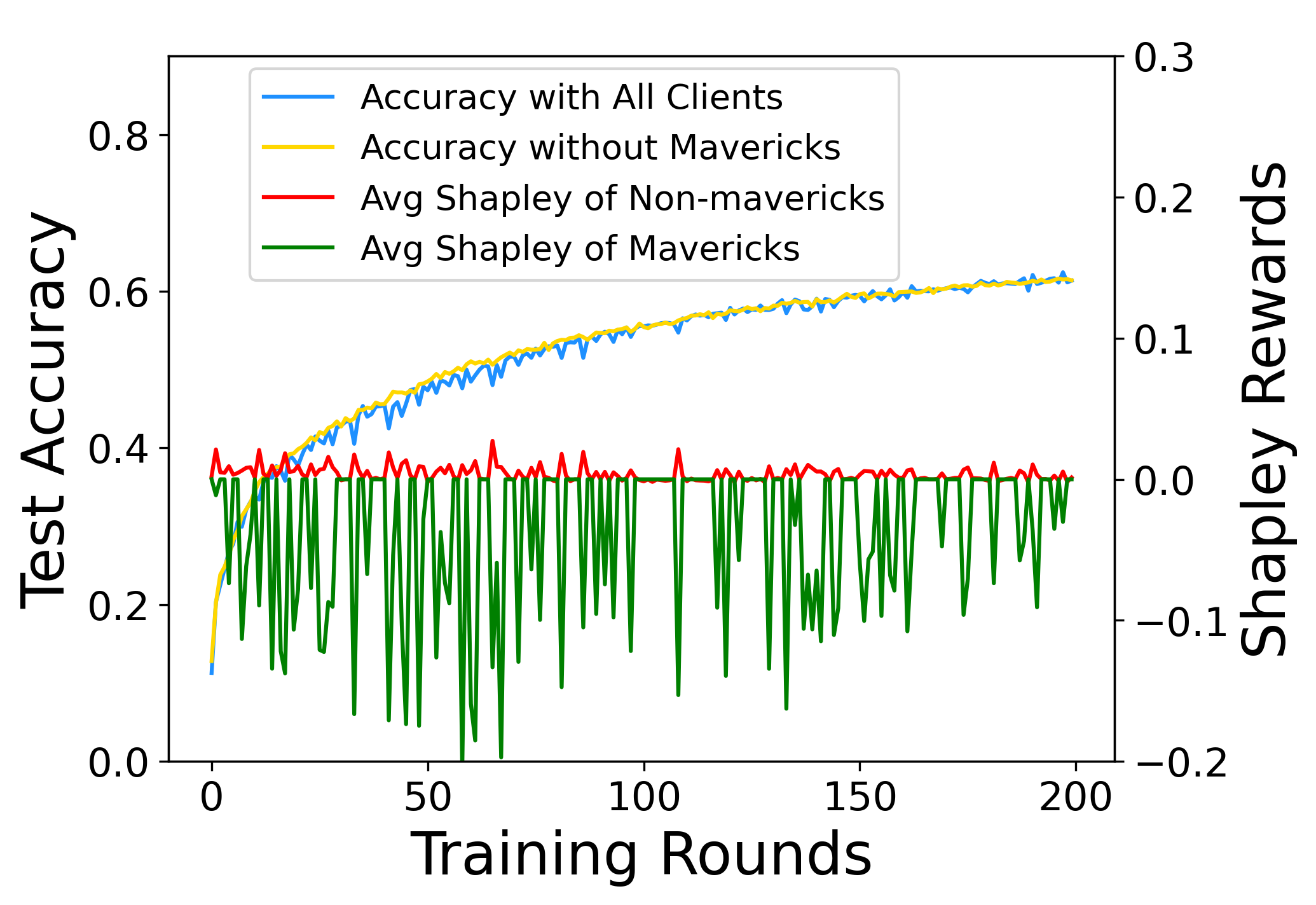}
\caption{S-FedAvg}
\label{fig:TMR_S_FedAvg}
    \end{subfigure}
    \hfill
    \begin{subfigure}{0.245\textwidth}
        \centering
        \includegraphics[width=\linewidth]{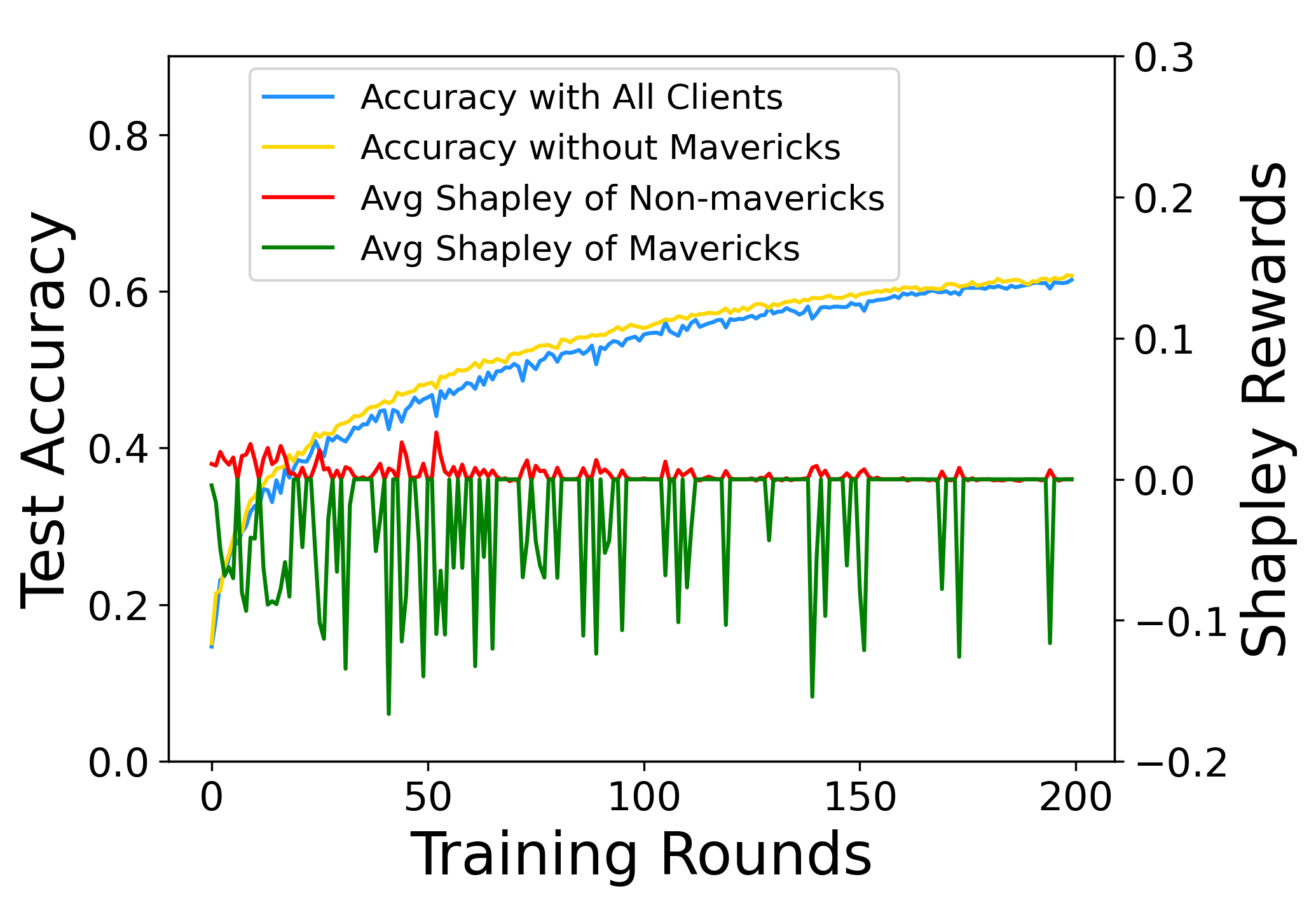}
\caption{FedEMD}
\label{fig:TMR_FedEMD}
    \end{subfigure}
    \centering
    \hfill
    \caption{Comparison of test accuracy and Shapley rewards with 50 clients (w/ client selection) for the CIFAR-10 dataset using TMR Shapley for various client selection techniques.}
    \label{fig:TMR_Cifar10}
\end{figure*}

In this section, we present additional experimental evaluations conducted on both MNIST and CIFAR-10 datasets concerning our reward and utility metrics. When examining the reward metrics, focusing on the fairer Shapley Rewards, we can observe from Figs.~\ref{fig:GTG_MNIST_appendix}, ~\ref{fig:GTG_Cifar10},~\ref{fig:MR_Cifar10}, and~\ref{fig:TMR_Cifar10} that FedMS provides more rewards to Mavericks compared to non-Mavericks for both MNIST and CIFAR-10 dataset, in line with the observed accuracy benefit of training with the \mave. In contrast, the state-of-the-art (SOTA) techniques provide lower rewards to Mavericks compared to non-Mavericks (example, FedAvg in Figs.~\ref{fig:GTG_FedAvg},~\ref{fig:MR_FedAvg},~\ref{fig:TMR_FedAvg} or FedEMD in Figs.~\ref{fig:GTG_FedEMD},~\ref{fig:MR_FedEMD},~\ref{fig:TMR_FedEMD}). Regarding the utility metric, we notice that FedMS better utilizes Mavericks, resulting in an overall improvement in model accuracy compared to the SOTA methods on both MNIST and CIFAR-10 datasets. From these two observations, we can deduce that our method not only effectively selects Mavericks, thereby enhancing model performance, but also ensures a fairer Shapley Rewards distribution
among \mave and non-\mave. Accuracy performance of the proposed \algoname in comparison with the SOTA baselines considering various SV approximation techniques for MNIST and CIFAR-10 dataset is given in Table~\ref{tab:Mnist_accuracy} and~\ref{tab:cifar10_accuracy}.

\begin{figure*}[t!]
    \centering
    \begin{subfigure}{0.245\textwidth}
        \centering
        \includegraphics[width=\linewidth]{images/cifar10/50_user_GTG_FedCSV.png}
\caption{FedMS (Our method)}
\label{fig:GTG_FedMS}
    \end{subfigure}
    \hfill
    \begin{subfigure}{0.245\textwidth}
        \centering
        \includegraphics[width=\linewidth]{{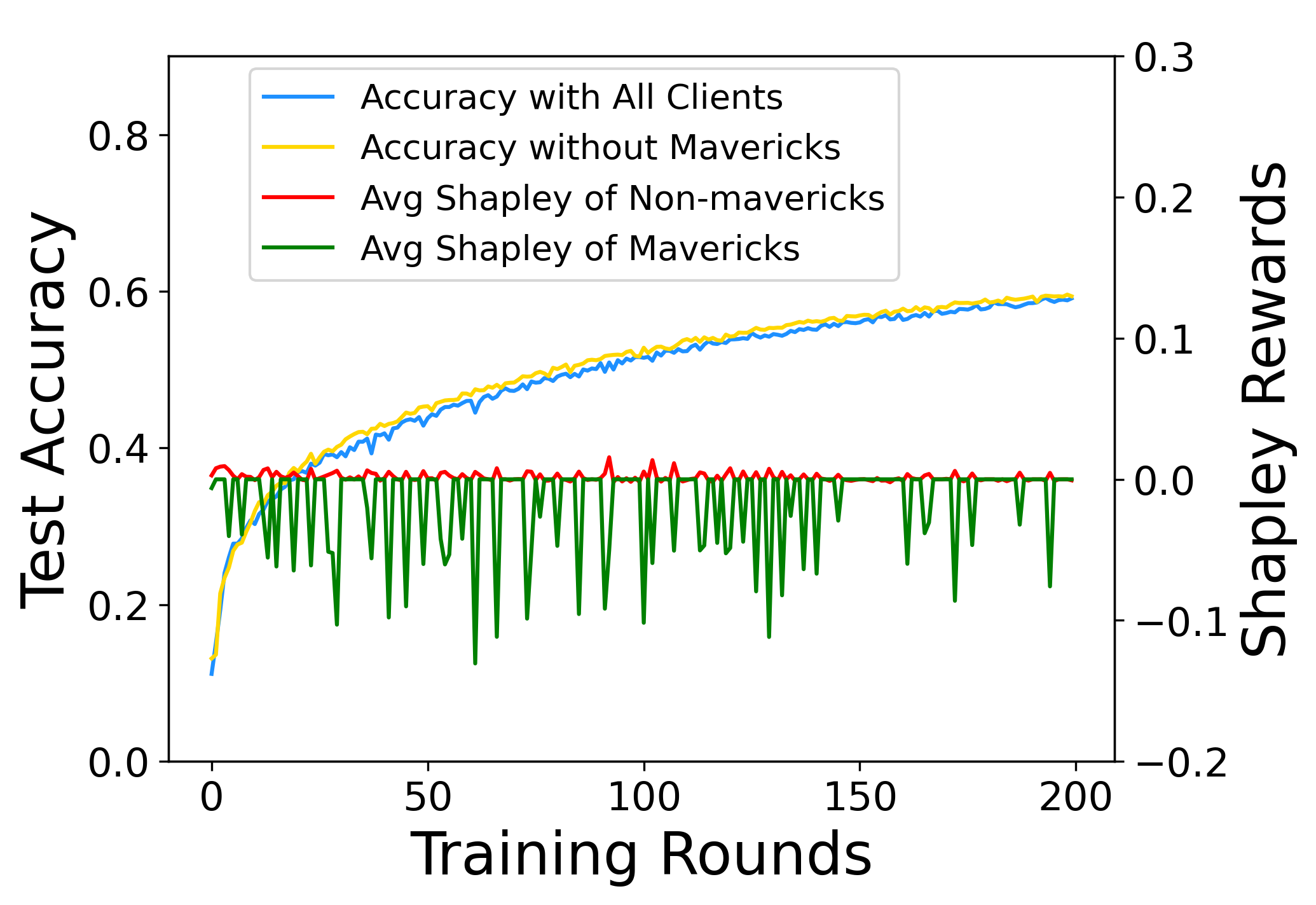}}
\caption{FedProx}
\label{fig:GTG_FedProx}
    \end{subfigure}
\begin{subfigure}{0.245\textwidth}
        \centering
        \includegraphics[width=\linewidth]{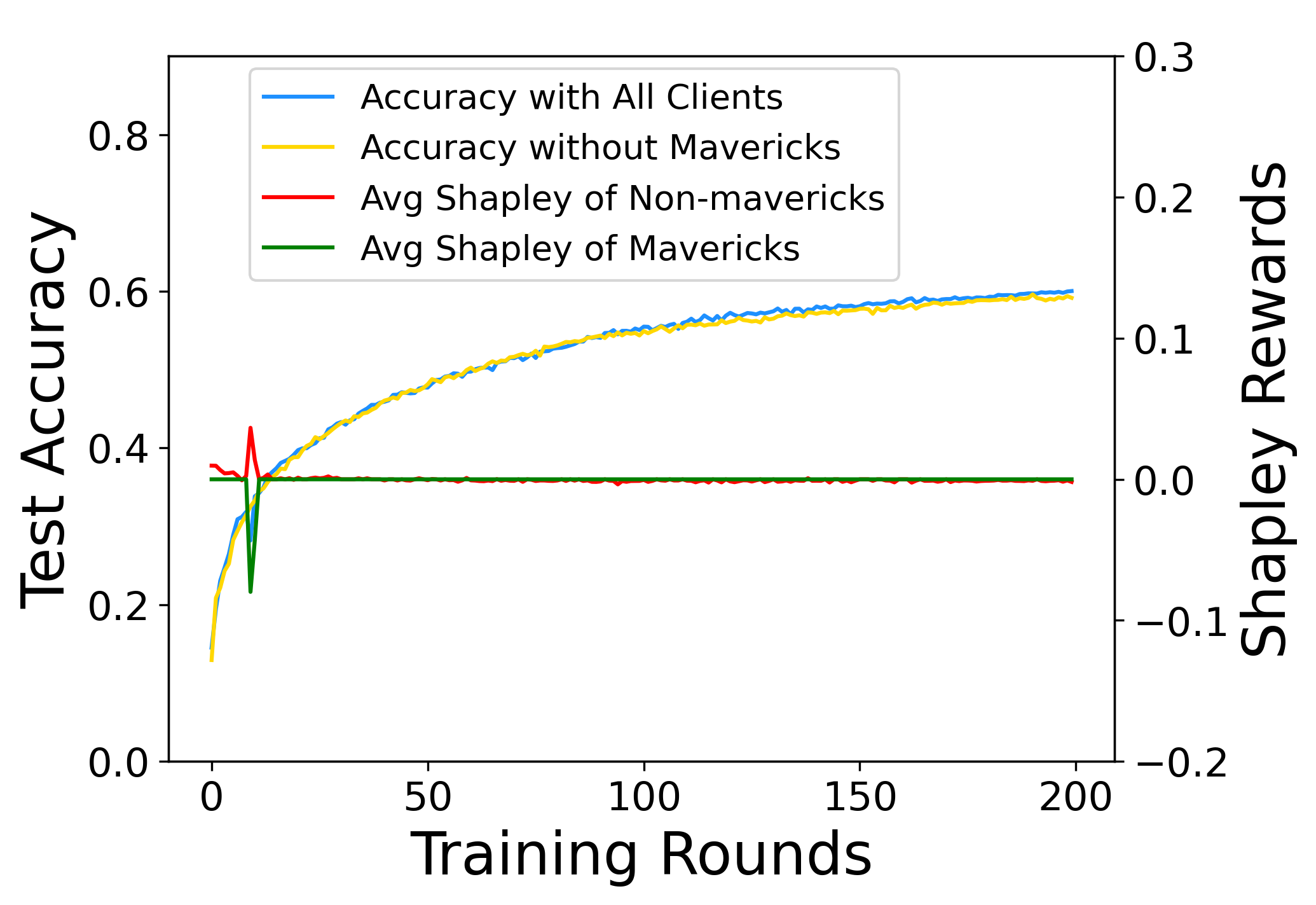}
\caption{GreedyFed}
\label{fig:GTG_Greedy}
    \end{subfigure}
    \hfill
\begin{subfigure}{0.245\textwidth}
        \centering
        \includegraphics[width=\linewidth]{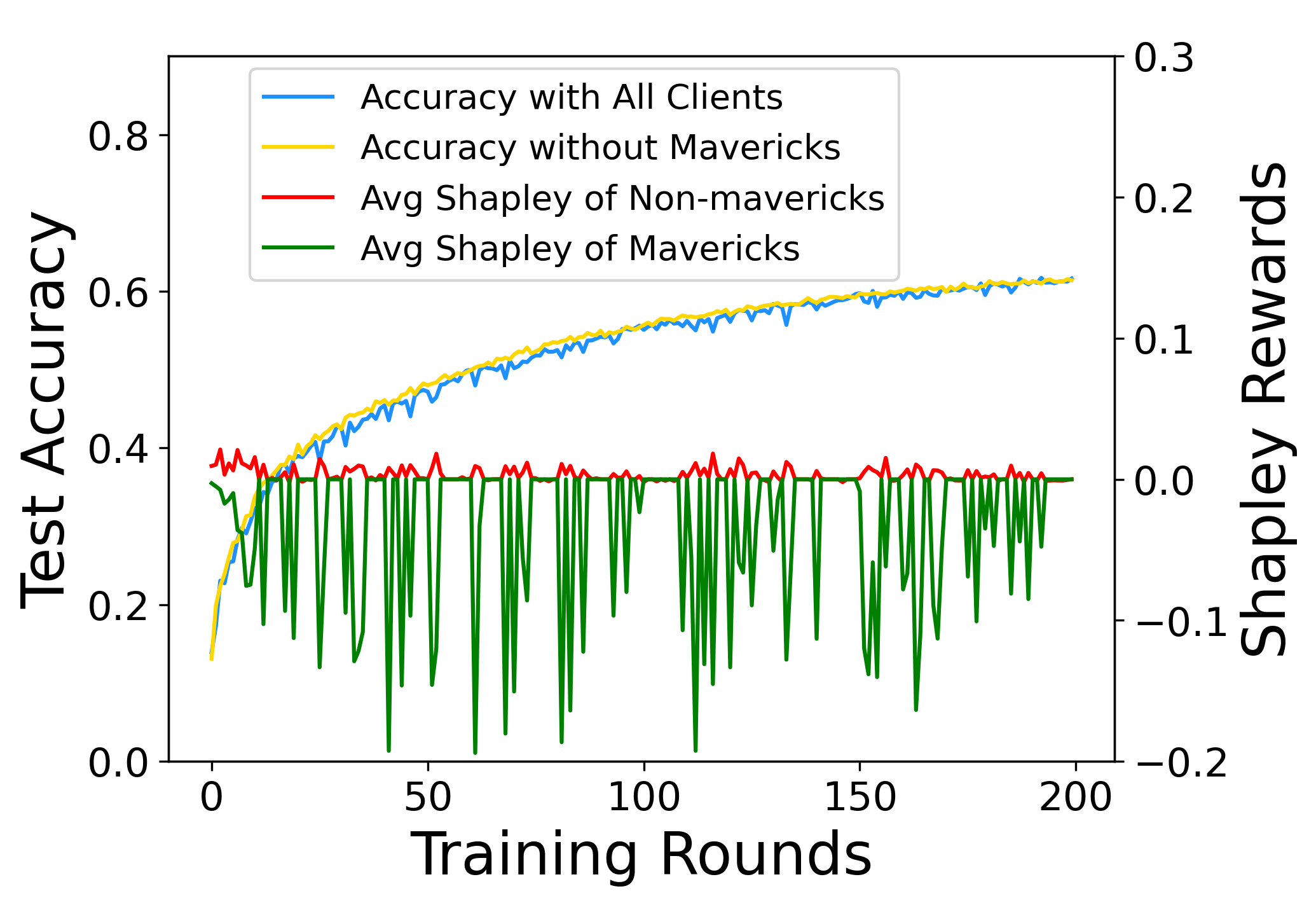}
\caption{PoC}
\label{fig:GTG_PoC}
    \end{subfigure}
    \hfill
    \caption{Comparison of test accuracy and Shapley rewards with 50 clients (w/ client selection) for the CIFAR-10 dataset using GTG-Shapley for various client selection techniques.}
    \label{fig:GTG_Cifar10}
\end{figure*}

\begin{figure*}[t!]
    \centering
    \begin{subfigure}{0.245\textwidth}
        \centering
        \includegraphics[width=\linewidth]{images/cifar10/50_user_MR_FedCSV.png}
\caption{FedMS (Our method)}
\label{fig:MR_FedMS}
    \end{subfigure}
    \hfill
    \begin{subfigure}{0.245\textwidth}
        \centering
        \includegraphics[width=\linewidth]{{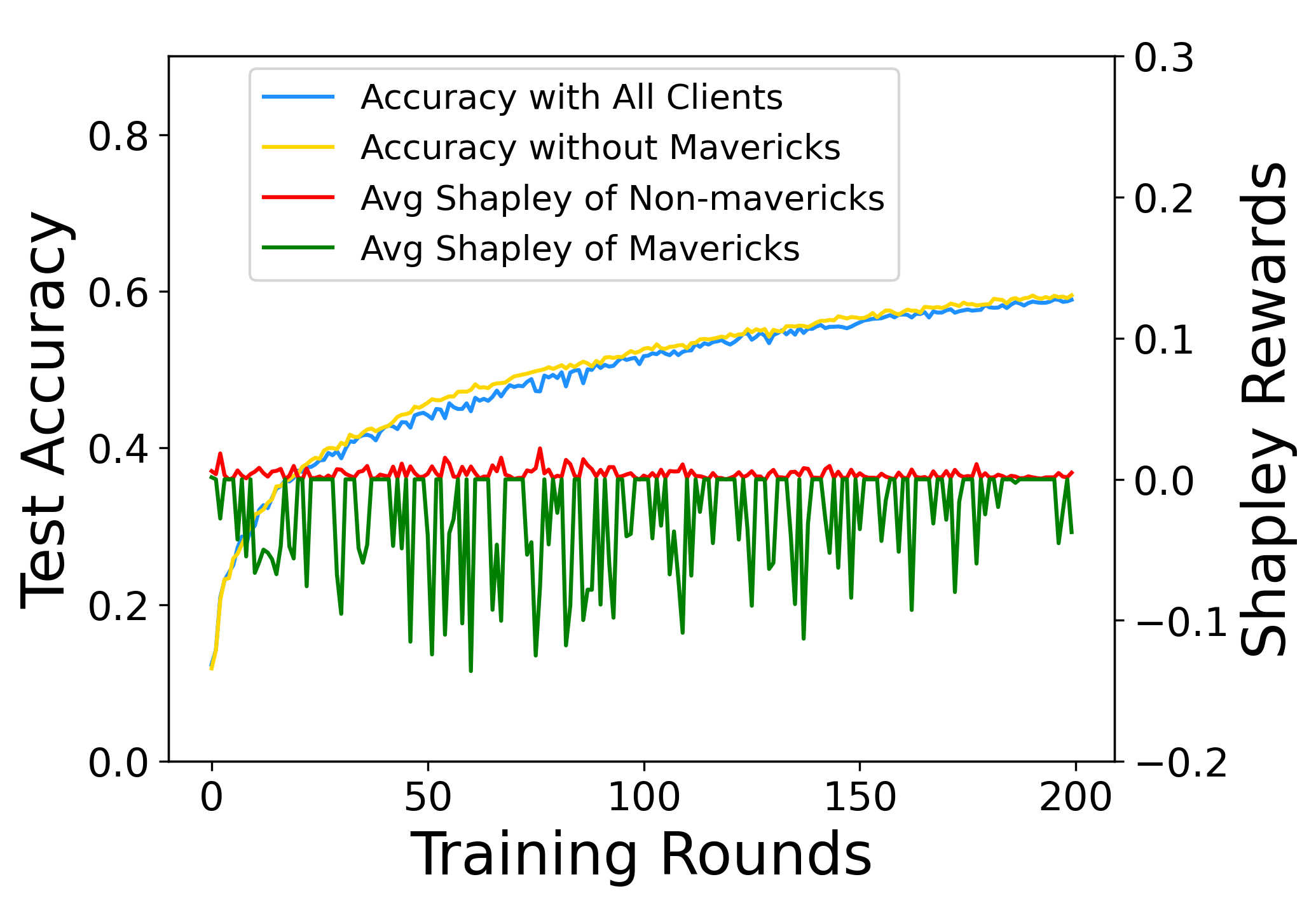}}
\caption{FedProx}
\label{fig:MR_FedProx}
    \end{subfigure}
\begin{subfigure}{0.245\textwidth}
        \centering
        \includegraphics[width=\linewidth]{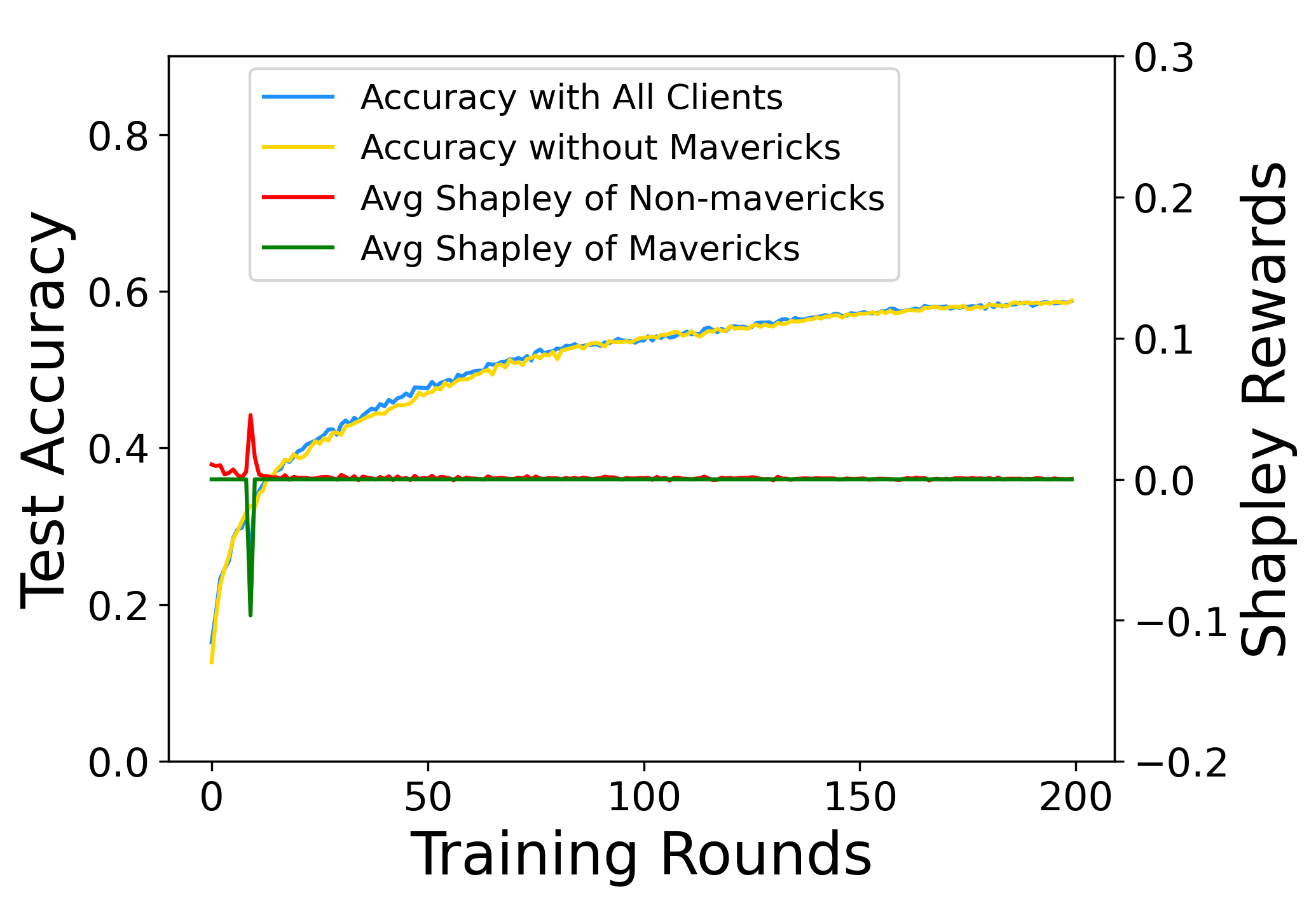}
\caption{GreedyFed}
\label{fig:MR_Greedy}
    \end{subfigure}
    \hfill
\begin{subfigure}{0.245\textwidth}
        \centering
        \includegraphics[width=\linewidth]{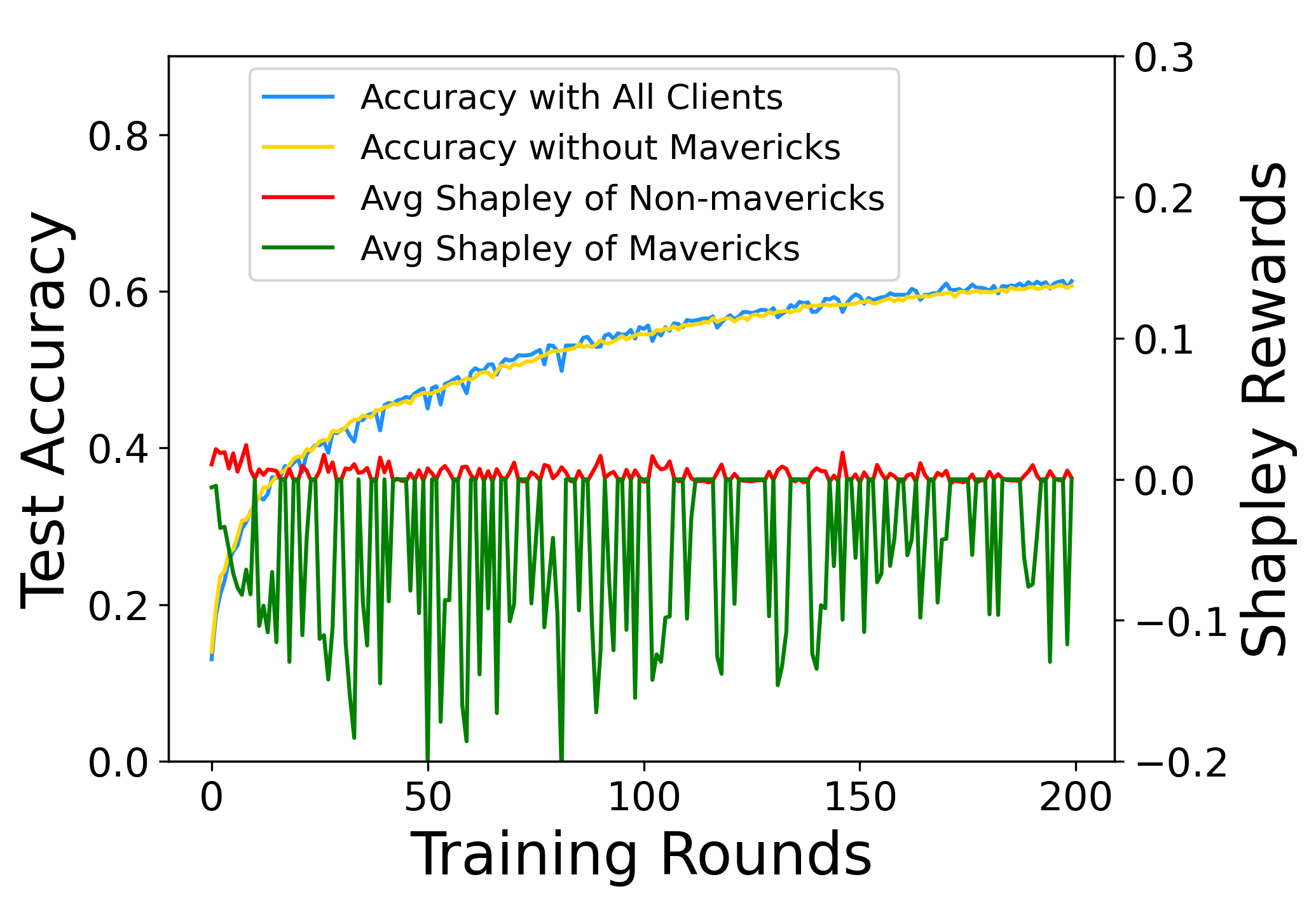}
\caption{PoC}
\label{fig:MR_PoC}
    \end{subfigure}
    \hfill
    \caption{Comparison of test accuracy and Shapley rewards with 50 clients (w/ client selection) for the CIFAR-10 dataset using MR Shapley for various client selection techniques.}
    \label{fig:MR_Cifar10}
\end{figure*}

\begin{figure*}[t!]
    \centering
    \begin{subfigure}{0.245\textwidth}
        \centering
        \includegraphics[width=\linewidth]{images/cifar10/50_user_TMR_FedCSV.png}
\caption{FedMS (Our method)}
\label{fig:TMR_FedMS}
    \end{subfigure}
    \hfill
    \begin{subfigure}{0.245\textwidth}
        \centering
        \includegraphics[width=\linewidth]{{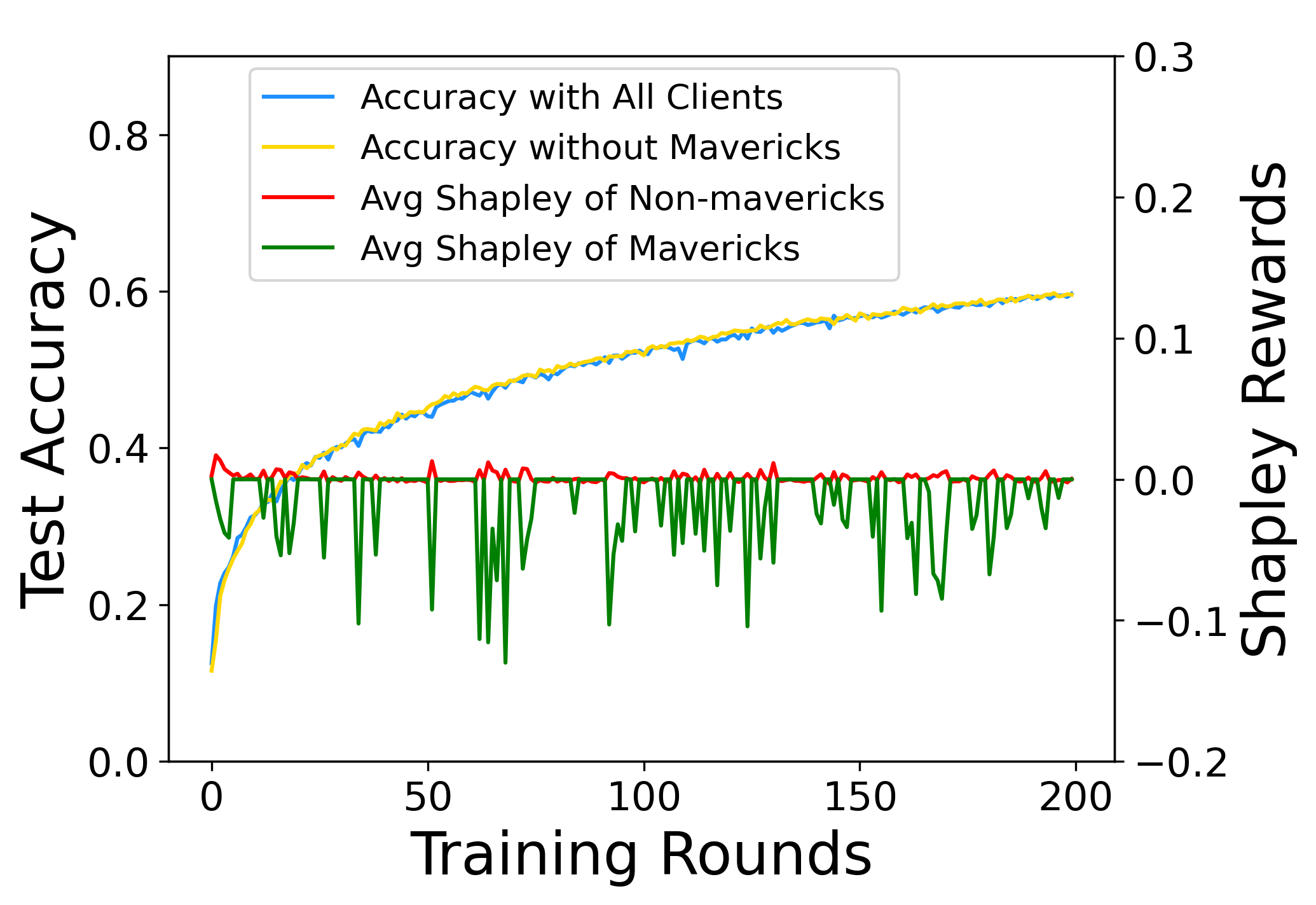}}
\caption{FedProx}
\label{fig:TMR_FedProx}
    \end{subfigure}
\begin{subfigure}{0.245\textwidth}
        \centering
        \includegraphics[width=\linewidth]{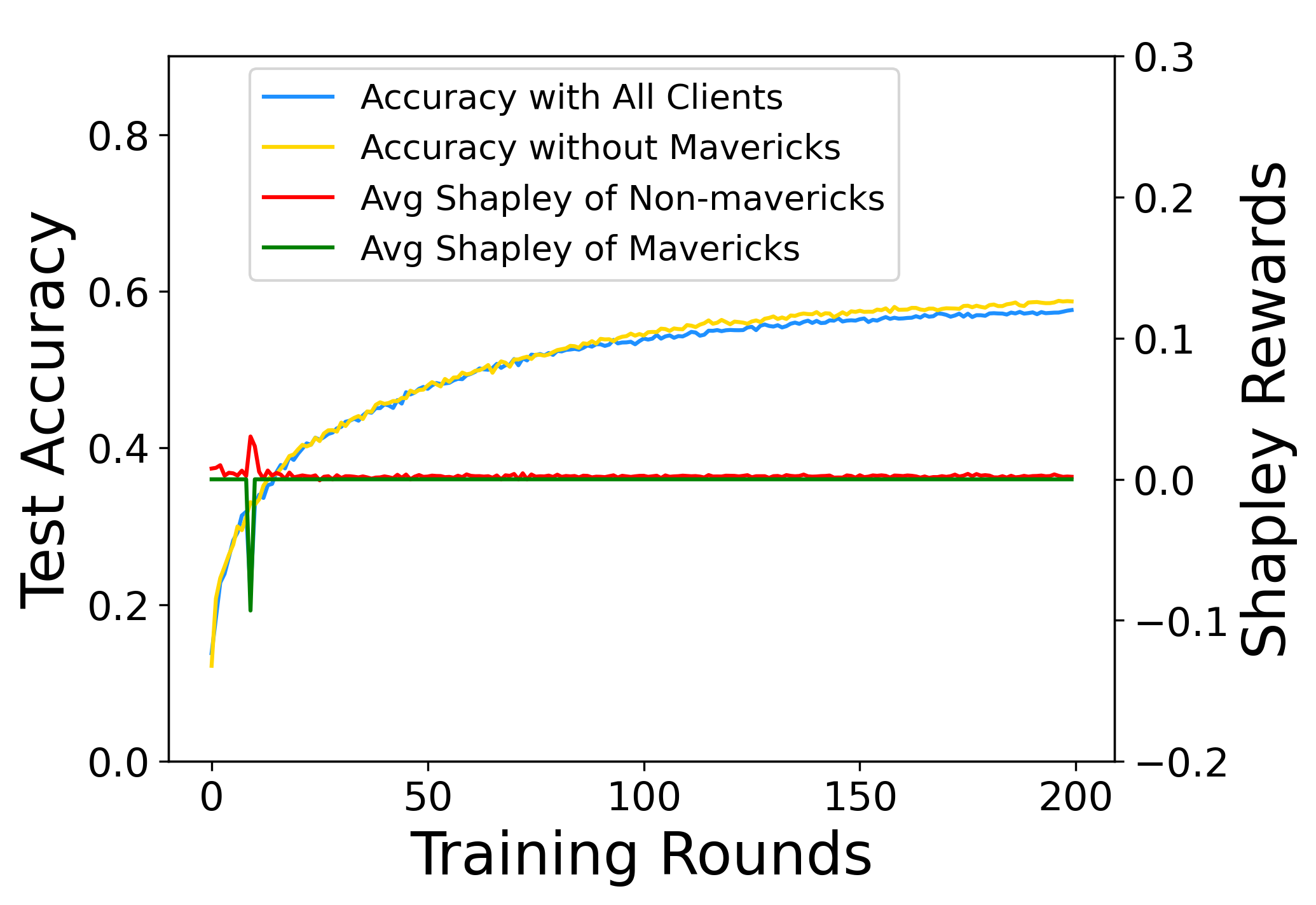}
\caption{GreedyFed}
\label{fig:TMR_Greedy}
    \end{subfigure}
    \hfill
\begin{subfigure}{0.245\textwidth}
        \centering
        \includegraphics[width=\linewidth]{images/cifar10/50_user_MR_PoC.png}
\caption{PoC}
\label{fig:TMR_PoC}
    \end{subfigure}
    \hfill
    \caption{Comparison of test accuracy and Shapley rewards with 50 clients (w/ client selection) for the CIFAR-10 dataset using TMR Shapley for various client selection techniques.}
    \label{fig:TMR_Cifar10}
\end{figure*}

\begin{table*}[ht!]
\centering
\begin{tabular}{|c|c|c|c|c|c|c|c|}
\hline
\diagbox[width=8em]{SHAP}{CS-Alg} & FedMS & FedAvg & FedProx & S-FedAvg & GreedyFed & PoC & FedEMD \\
\hline
GTG-Shapley & \textbf{82.91} $\pm$ 0.5 & 73.77 $\pm$ 0.1  & 73.52 $\pm$ 0.4 & 74.00 $\pm$ 0.5 & 73.91 $\pm$ 0.2 & 74.49 $\pm$ 0.7& 73.79 $\pm$ 0.2\\
\hline
MR & \textbf{82.81} $\pm$ 1.8 &  73.99  $\pm$ 0.4 & 73.79 $\pm$ 0.4 & 73.94 $\pm$ 0.2 & 73.66 $\pm$ 0.2 & 73.94 $\pm$ 0.1 & 73.87 $\pm$ 0.1\\
\hline
TMR & \textbf{80.27} $\pm$ 2.4 & 73.74  $\pm$ 0.1 & 74.02 $\pm$ 0.4 & 73.67 $\pm$ 0.1 & 73.99 $\pm$ 0.1 & 73.42 $\pm$ 0.4 & 75.49 $\pm$ 1.0\\
\hline
\end{tabular}
\caption{Model performance (test accuracy in \%) of different client selection algorithms (CS-Alg) including FedMS for MNIST dataset under various Shapley value approximation methods.}
\label{tab:Mnist_accuracy}
\end{table*}

\begin{table*}[ht!]
\centering
\begin{tabular}{|c|c|c|c|c|c|c|c|}
\hline
\diagbox[width=8em]{SHAP}{CS-Alg} & FedMS & FedAvg & FedProx & S-FedAvg & GreedyFed & PoC & FedEMD \\
\hline
GTG-Shapley & \textbf{64.79} $\pm$ 0.5  & 60.87 $\pm$ 0.1   & 60.25 $\pm$ 0.4 & 61.53 $\pm$ 0.3 & 59.87 $\pm$ 0.1 & 61.65 $\pm$ 0.3 & 62.7 $\pm$ 0.1 \\
\hline
MR & \textbf{64.56} $\pm$ 1.5 & 61.84 $\pm$ 0.2 & 60.97 $\pm$ 0.4 & 61.24 $\pm$ 0.2 & 58.81 $\pm$ 0.2 & 62.29 $\pm$ 0.3 & 62.25 $\pm$ 0.1 \\
\hline
TMR & \textbf{64.5} $\pm$ 0.6 & 61.84 $\pm$ 0.1 & 59.9 $\pm$ 0.1 & 61.5 $\pm$ 0.5 & 57.7 $\pm$ 0.2 & 61.25 $\pm$ 0.2 & 61.85 $\pm$ 0.15 \\
\hline
\end{tabular}
\caption{Model performance (test accuracy in \%) of different client selection algorithms (CS-Alg) including FedMS for CIFAR-10 dataset under various Shapley value approximation methods.}
\label{tab:cifar10_accuracy}
\end{table*}

\end{document}